\documentclass{article}

\usepackage{arxiv}

\usepackage[utf8]{inputenc} 
\usepackage[T1]{fontenc}    
\usepackage{hyperref}       
\usepackage{url}            
\usepackage{booktabs}       
\usepackage{amsfonts}       
\usepackage{nicefrac}       
\usepackage{microtype}      
\usepackage{lipsum}		
\usepackage{hyperref}
\usepackage{array}
\usepackage{subfig}
\usepackage{amsmath}
\usepackage{bbm}
\usepackage{tikz}
\usepackage{siunitx}
\usepackage{algorithm}
\usepackage{algpseudocode}
\usepackage{todonotes}

\algnewcommand{\LineComment}[1]{\State \(\triangleright\) #1}
\newcolumntype{C}[1]{>{\centering\arraybackslash}m{#1}}

\usepackage{mathrsfs} 
\usetikzlibrary{3d,decorations.markings,calc}
\usepackage{tikz-3dplot}
\makeatletter 
\tikzoption{canvas is xy plane at z}[]{%
  \def\tikz@plane@origin{\pgfpointxyz{0}{0}{#1}}%
  \def\tikz@plane@x{\pgfpointxyz{1}{0}{#1}}%
  \def\tikz@plane@y{\pgfpointxyz{0}{1}{#1}}%
  \tikz@canvas@is@plane
}

\title{\LARGE \bf
QED: using Quality-Environment-Diversity to evolve resilient robot swarms\\
}
\author{David M. Bossens and Danesh Tarapore$^{}$
\thanks{$^{}$Authors are with the School of Electronics and Computer Science, University of Southampton, SO17 1BJ Southampton, U.K.
        {\tt\small d.m.bossens@soton.ac.uk}}%
}

\begin{document}

\maketitle
\thispagestyle{empty}
\pagestyle{empty}


\begin{abstract} 
In swarm robotics, any of the robots in a swarm may be affected by different faults, resulting in significant performance declines. To allow fault recovery from randomly injected faults to different robots in a swarm, a model-free approach may be preferable due to the accumulation of faults in models and the difficulty to predict the behaviour of neighbouring robots. One model-free approach to fault recovery involves two phases: during simulation, a quality-diversity algorithm evolves a behaviourally diverse archive of controllers; during the target application,  a search for the best controller is initiated after fault injection. In quality-diversity algorithms, the choice of the behavioural descriptor is a key design choice that determines the quality of the evolved archives, and therefore the fault recovery performance. Although the environment is an important determinant of behaviour, the impact of \textit{environmental diversity} is often ignored in the choice of a suitable behavioural descriptor. This study compares different behavioural descriptors, including two generic descriptors that work on a wide range of tasks, one hand-coded descriptor which fits the domain of interest, and one novel type of descriptor based on \textit{environmental diversity}, which we call Quality-Environment-Diversity (QED). Results demonstrate that the above-mentioned model-free approach to fault recovery is feasible in the context of swarm robotics, reducing the fault impact by a factor 2-3. Further, the environmental diversity obtained with QED yields a unique behavioural diversity profile that allows it to recover from high-impact faults. 
\end{abstract}


%

\maketitle

\section{Introduction}
Swarm robotics \cite{Sahin2005,Brambilla2013a} studies the emergence of collective behaviour in large-scale teams of robots. The robots in a swarm are simple and have limited sensory, computational and communication capabilities. The tasks they have to accomplish are relatively complex, and may not be achievable by individual members of the swarm \cite{Bayindir2016}. Robot swarms have been investigated in numerous studies, to perform loosely-coordinated tasks such as collaborative exploration, monitoring and surveillance \cite{Schwager2007,Duarte2016}, as well as tightly-coordinated tasks such as self-assembly \cite{Garattoni2018}, coordinated movement \cite{Viragh2014} and foraging \cite{Gross2009,llenas2018quality}.

Despite the robustness conferred to robot swarms by the decentralised nature of inter-robot coordination \cite{Bonabeau1999}, they remain brittle systems that are rendered inoperable when inadvertent faults are sustained by individual robots of the swarm. Studies on fault tolerance in swarm robotics have revealed that even partial failures to one or a few robots may significantly hamper the capability of the swarm to complete its mission \cite{Bjerknes,Winfield2006}. In developing fault-tolerant robot swarms, many studies have investigated fault-detection algorithms, for individual robots of the swarm to robustly detect faults, both endogenously in themselves \cite{Tarapore2015,Christensen2008}, and exogenously in neighboring robots \cite{millard2016exogenous,Christensen2009,Tarapore2017}. However, to the best of our knowledge, fault-recovery in swarm robotics, wherein robots of the swarm adapt their behaviour to compensate for the different faults that they may sustain, is an open challenge \cite{Yang2018}.

Fault recovery has been previously investigated in the context of single-robot systems with multiple actuated degrees of freedom providing redundancy, such as quadruped and hexapod walking robots and multi-joint pick-and-place robot arms \cite{chatzilygeroudis2018survey}. Many of these studies are model-based, and involve updating the model of the robot to restore the accuracy of movements when unexpected faults, such as damages in the actuators, perturb the robot-environment interaction \cite{Bongard2006,Koos2013b,Chatzilygeroudis2018a}. While such an approach is promising when the model learned is accurate, its extension to swarm robotic systems seems elusive, consequent to the following: a) even small deviations of the self-model to reality may accumulate rapidly when considering the large number of robots in the swarm; and b) the self-model of one robot may not be able to anticipate changes due to faults sustained by other robots of the swarm. An alternative approach, explored for a hexapod robot with damaged actuators \cite{Cully2015}, is to recover from faults by intelligently searching over a diverse archive of walking behaviours evolved using \textit{quality-diversity algorithms} \cite{Pugh2016a,Cully2018}, without the need for a model. Therefore, such a model-free approach appears promising for fault recovery in a robot swarm. 

An important aspect of quality-diversity algorithms is the choice of behavioural descriptor used to characterise the behaviours of the evolved solutions. Many studies employing quality-diversity algorithms have relied on hand-coded descriptors based on domain-specific insights or dimensions which are of particular interest to the end-user \cite{Mouret2015,Pugh2015,Nordmoen2018,Engebraten,Hart2018}. Generic behavioural descriptors have recently been proposed, including methods which automatically derive the behavioural description from the sensor-actuator trajectories of the robot \cite{Meyerson2016,Gomez2009}, such as Stochastic Policy Induction for Relating Inter-task Trajectories (SPIRIT) \cite{Meyerson2016}, without the need for any domain-specific information. Other approaches such as Systematically Derived Behaviour Characterisations (SDBC) \cite{Gomes2014} exploit domain-specific information in a systematic manner, deriving the behavioural description from the averaged relations between different entities, such as robots and objects of interest in the environment.
Importantly, for swarm robotic systems, the effect of the choice of behavioural descriptor on the quality of the evolved archive \textit{for fault recovery} remains to be investigated.

A commonality amongst quality-diversity algorithms is that all of the individuals in the archive are evaluated in the same operating environment. However, the bias-variance dilemma \cite{GemanS.BienenstockE.&Doursat} implies that controllers trained or evolved in a specific environment will excel in that environment but may fail to generalise to the larger domain of interest.
To allow the evolution of robust robot controllers, various studies have explored modifying the fitness evaluation to provide tolerance to faults injected in the robot hardware \cite{Thompson1999}, to bridge the robot simulation-reality gap \cite{Koos2013c,IlgeAkkaya}, and produce robot controllers with better generalization capabilities \cite{Pinville2011}. Moreover, recent work in active curriculum learning \cite{Schmidhuber2013,Srivastava2012} and open-ended co-evolution \cite{Wang2019,Brant2017} demonstrate the beneficial impact of enabling controllers to learn on ever-more challenging and diverse environments targeted to the controller's current skill levels. Other studies in evolutionary computation show that recording a variety of solutions associated with the task-objective solved, provides evolution with stepping stones leading to higher behavioural diversity and the ability to solve problems of greater complexity \cite{Nguyen2016,Huizinga2018}. These findings demonstrate that some robot behaviours may best, if not \textit{only}, be evolved by allowing learning in a number of different environments.

We investigate the effect of incorporating environmental diversity in quality-diversity algorithms on the quality of evolved archive of solutions, and consequently on the fault-recovery performance of the robot swarm. Quality-diversity algorithms may be improved by evaluating robot swarms in a diversity of environments because the selected behavioural descriptors may omit important information on the dynamics of the swarm (e.g., the summary statistics of the SDBC descriptors \cite{Gomes2014} may fail to capture intricacies of the inter-robot interactions at finer temporal resolutions). Additionally, a given behaviour may be more easily characterised by the type of environment in which it is high-performing (e.g., a high-performing behaviour for patrolling \textit{a cluttered arena with a low robot density swarm}). We formulate a novel framework called Quality-Environment Diversity (QED) which evaluates individuals in the archive in different environments and characterises behaviour using a description of the environment. The QED framework is evaluated by analysing the behavioural diversity of the evolved solutions, as well as their ability to recover from a variety of eight different sensor/actuator faults injected at random to each of the robots in a swarm. Our paper makes two important contributions: a) it provides a comparative study between a hand-coded behavioural descriptor, and the SPIRIT \cite{Meyerson2016} and SDBC \cite{Gomes2014} generic behavioural descriptors, to assess the impact of the choice of descriptor on fault recovery; and b) it introduces the proposed QED algorithm, which compared to the hand-coded and generic behavioural descriptors, demonstrates higher behavioural diversity and performance in fault recovery, across five commonly employed benchmark swarm robotic tasks. 



\section{Quality-environment diversity algorithm}
The QED algorithm builds a diverse archive of robot swarm behaviours by evaluating the robot swarm controllers in a repertoire of different randomly selected environments centred around a normal operating environment $\mathcal{E}$, which simulates the physical environment of the robot swarm. Our implementation of QED is based on Multi-dimensional Archive of Phenotypic Elites (MAP-Elites) \cite{Mouret2015}, a quality-diversity algorithm used in numerous studies in evolutionary computation \cite{Vassiliades2018c,TaraporeClune2016,Hart2018,Mouret2015a,Engebraten,Mouret2015,Cully2018,Nordmoen2018}. With MAP-Elites, a topologically organised behaviour-performance map $\mathcal{M}$ is evolved. The dimensions of variation of the map $j \in \{1,\dots,D\}$ are defined a priori and correspond to the behavioural descriptors characterising the robot swarm's behaviour. Each of the $D$ dimensions of the map are discretised following user-requirements or memory storage constraints. Given the discretisation, the MAP-Elites algorithm searches for the highest performing solution for each cell in the map.

\paragraph{Quality-environment diversity with MAP-Elites.} 

Our QED algorithm differs from MAP-Elites in that solutions are located in the map $\mathcal{M}$ based on the environment they were evaluated in using an \textit{environment descriptor} -- rather than their behavioural descriptor. The QED algorithm is initialised with an empty map $\mathcal{M}$, and first generates a set of random controllers $P$. An environment generator then randomly generates an environment $\tilde{\mathcal{E}}$, one for each controller $i \in \{1,\dots,|P|\}$, and evaluates the performance of the controller in that environment. Each of the evaluated solutions is described by the environment descriptor of the environment it was evaluated in. Finally, similar to MAP-Elites, if the performance of the evaluated solution exceeds that of the current solution at that location in the environment-performance map, it is added to the map, replacing the solution at that location. Therefore, solutions are only retained in the environment-performance map if they are the best for the environment type defined by their location in the environment-performance map or if no solution exists at that location.

Upon completion of this initialisation step, the QED improves the solutions in the environment-performance map through: i) the generation of new environments; and ii) random variation and selection of the existing solutions in the map. At each iteration, the algorithm picks a solution from the map at random, following a uniform distribution. A copy of that solution is then randomly mutated (see Table~\ref{tab:evolution-parameters} for parameters of mutation operators). The environment generator generates a random environment $\tilde{\mathcal{E}}$ for the mutated solution, which determines its location in the environment-performance map. The mutated solution is evaluated in $\tilde{\mathcal{E}}$, and is retained if it outperforms the current solution at that location in the environment-performance map. The evolutionary process is repeated until the maximal number of evaluations is expended. An implementation of the QED algorithm is illustrated in Algorithm~\ref{alg:QED-MAPElites}.

\begin{algorithm}
  \caption{Quality-Environment Diversity algorithm implemented with MAP-Elites, to generate an environment-performance map with dimensionality $D$.}
  \label{alg:QED-MAPElites}
    \begin{algorithmic}[1] 
      \State $\mathcal{M} \gets \emptyset$ \Comment{Creation of an empty $D$-dimensional map.}
      \For{$i=1$ to $p$} \Comment{Generate a random initial population of size $p$ (we choose $p=2000$).}
      \State $P[i] \gets \texttt{random-controller()}$ \Comment{Randomly create neural-network controller (see Table~\ref{tab:evolution-parameters} for details).}
      \State Perform \texttt{add-controller}($P[i]$)
      \EndFor

      \For{$i=1$ to $I$ } \Comment{Repeat for $I$ iterations.}
      \State $c \gets \texttt{select-random}(\mathcal{M})$ \Comment{Sample controller $c$ from uniform distribution over non-empty cells in $\mathcal{M}$.}
      \State $c' \gets \texttt{mutate}(c)$ \Comment{For parameters of mutation operator see Table~\ref{tab:evolution-parameters}.}
      \State Perform \texttt{add-controller}($c'$)
      \EndFor 

      \Procedure{add-controller}{controller $c$}
      \State Randomly select $A_j \in \mathbf{P}_j \mbox{ } \forall j \in \{1,\dots,D\}$
      \State Generate environment $\tilde{\mathcal{E}}$ parametrised by $\mathbf{A} = \langle A_1 \dotsc A_D\rangle$


      \State $\beta \gets \texttt{environment-descriptor}(\tilde{\mathcal{E}})$ \Comment{Describe $c$ by the environment $\tilde{\mathcal{E}}$ it is to be evaluated in.}
      \State Evaluate $f(\tilde{\mathcal{E}},c)$ \Comment{Evaluate controller $c$ in environment $\tilde{\mathcal{E}}$.}

      \If {$\mathcal{M}[\beta] = \emptyset$ \textbf{ or } $f(\tilde{\mathcal{E}},c) > f(\tilde{\mathcal{E}},\mathcal{M}[\beta])$}
      \State $\mathcal{M}[\beta] = c$ \Comment{Add individual $c$ to the map $\mathcal{M}$.}      
      \EndIf
      \EndProcedure
    \end{algorithmic}
\end{algorithm}

\begin{table}
  \caption{Environment attributes of the QED algorithm, and their value in normal and perturbed environments. The environment attributes are characteristics of the robots of the swarm, and their operating environment.}
  \label{tab:attributeVec}
  \begin{tabular}{p{2cm} p{5.5cm} p{2.5cm} l}
    \textbf{Environment Attribute} & \textbf{Description}  &  \textbf{Value in normal environment}& \textbf{Range of perturbations injected} \\ \hline
    $A_1$ & Robots' maximal linear speed & $\SI{10}{cm/s}$ & $\mathbf{P}_1=\{5,10,15,20\}\SI{}{cm/s}$ \\
    $A_2$ & Number of robots in the swarm & $10$ & $\mathbf{P}_2=\{5,10,15,20\}$  \\
    $A_3$ & Arena size & $\SI{16}{m^2}$ & $\mathbf{P}_3=\{4,9,16,25\}\SI{}{m^2}$  \\
    $A_4$ & Number of obstacles & $0$ & $\mathbf{P}_4=\{0,2,4,6\}$ \\
    $A_5$ & Robots' range-and-bearing sensor range & $\SI{1}{m}$ & $\mathbf{P}_5=\{25,50,100,200\}\SI{}{cm}$\\
    $A_6$ & Robots' proximity sensor range & $\SI{11}{cm}$ & $\mathbf{P}_6=\{5.5,11,22,44\}\SI{}{cm}$
  \end{tabular}
\end{table}

\paragraph{Environment generation for quality-environment diversity.} The generation of a diverse set of environments is an essential aspect of the QED algorithm. In our implementation of the QED, the environment for the robot swarm is characterised by the following six attributes $A = \langle A_1, \dotsc, A_6\rangle$: i) maximum linear speed of the robots in the swarm; ii) size of the robot swarm; iii) size of the arena the swarm is operating in; iv) number of obstacles in the arena; and v) maximum range-and-bearing sensor range of the robots in the swarm; and vi) maximum proximity sensor range of the robots in the swarm. The environment attributes are selected to elicit a diverse repertoire of swarm behaviours from perturbations reshaping the pathways of robots' sensory-motor interactions. In our environment generator, the values for each of the attributes $A_j, \mbox{ } j \in \{1, \dots, 6\}$ of the generated environments are randomly selected following a uniform distribution from a select set of perturbations around the value of the attribute in a normal operating environment. The selected perturbations on the environment attributes typically varied in range from a quarter of to two-fold the normal value of the attribute (see Table~\ref{tab:attributeVec} for details on the range of perturbations).



\section{Experimental method}
\subsection{Swarm robot platform}
\label{sec:robdetails}
\begin{figure}
\centering
\includegraphics[width=0.25\textwidth]{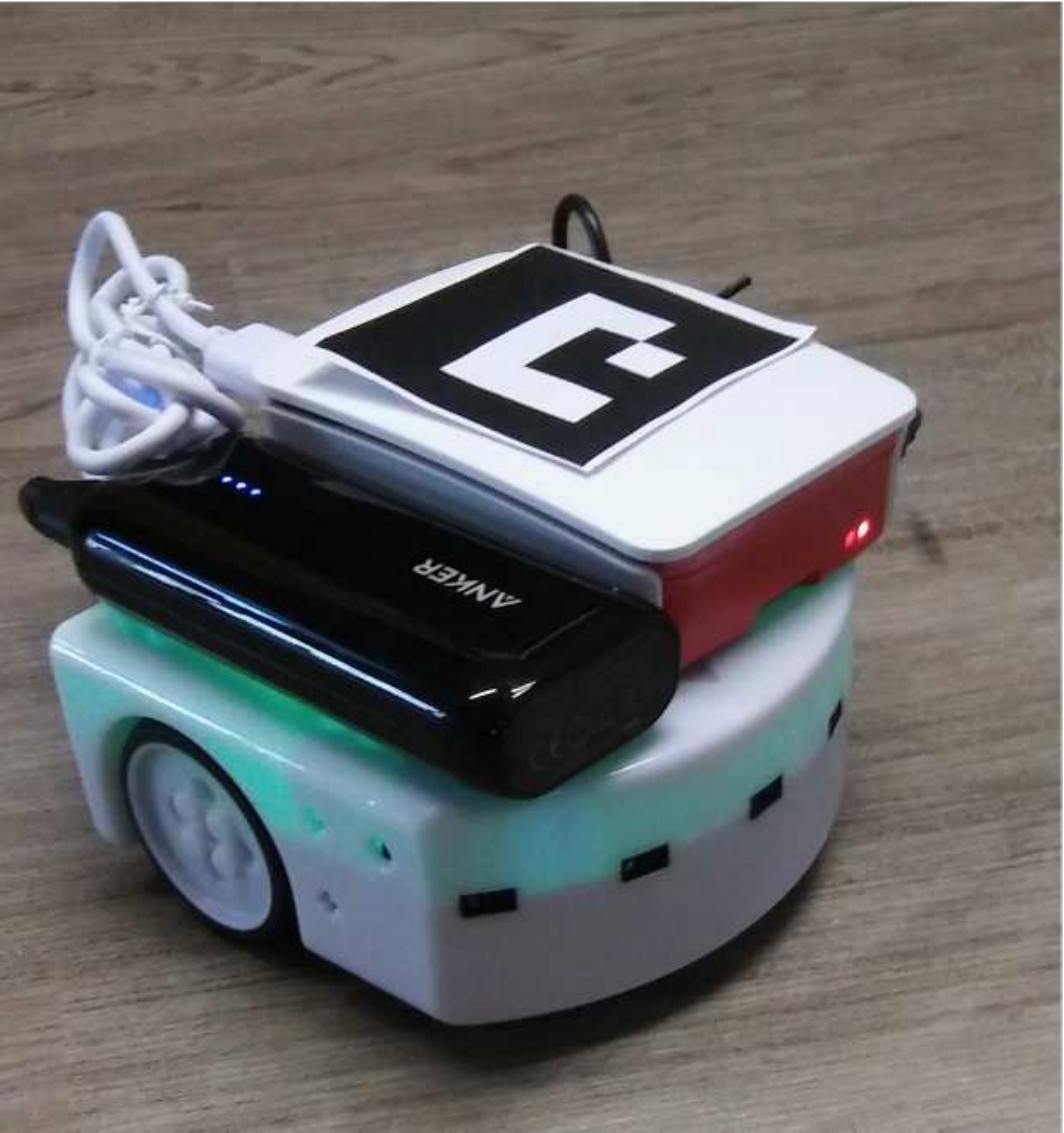}
 \caption{Thymio II robot augmented with a Raspberry Pi 3 B+ model.} \label{fig: thymio}
\end{figure}


We use a physics-based, discrete-time robot swarm simulator named ARGoS \cite{Pinciroli2012}, designed to realistically simulate complex swarm behaviours. The robot swarm simulated is composed of 10 two-wheel differential-drive Thymio mobile robots (\cite{Riedo2013}) in an environment of size $4\times4~\SI{}{m^2}$ (see Fig. \ref{fig: thymio}). The Thymio robot is \SI{11}{cm} long and \SI{8.5}{cm} wide, with a maximum linear speed of \SI{10}{cm/s}, a maximum angular speed of $\SI{127.32}{^\circ/s}$, and a control cycle of \SI{0.20}{s}. The Thymio robot model is equipped with five frontal and two rear infrared proximity sensors of range \SI{11}{cm} for obstacle avoidance and two actuators which control the robot's speed and direction. To localise neighbouring robots in the swarm, the experiments augment each of the Thymio robots of the swarm with range-and-bearing sensors, which have a maximal range of \SI{1}{m} and which discretise the bearing angles into eight cones each of size $\SI{45}{^\circ}$.

\begin{table}
\centering
\caption{Parameters for evolution of neural network controllers for the robot swarm.}
\label{tab:evolution-parameters}
\begin{tabular}{l | l}
\textbf{Parameter} & \textbf{Values} \\ 
\hline
Network initialisation & Random topology and weights \\
Number of neurons & $\{0,\dots,20\}$\\
Number of connections & $\{0,\dots,40\}$\\
Weight range of neural connections & $[-2,2]$\\
Neuron activation function & $\tanh$\\
Neuron addition \& deletion rate & $0.10$ each \\
Connection addition, deletion, \& modification rate & $0.15$ each \\
Connection modification type & random change of a connection's incoming or outgoing neuron \\
Weight mutation rate & $0.05$ \\
Weight mutation type & Polynomial mutation with $\eta_m=15$ (see \cite{Deb2001})\\
Initial number of solutions & $2000$ \\
Generations of evolution (solutions evaluated per generation) & $30000$ ($80$)\\ 
\end{tabular}  
\end{table}

The robot swarms in our experiments are homogeneous, i.e., each and every robot in the swarm is controlled by clonal copies of the same neural network controller. For each robot, the neural network takes as inputs the activations of the 7 proximity sensors, as well as 8 range-and-bearing sensor inputs, and a bias activation. The range-and-bearing sensors provide as input the range of the closest neighbouring robot with bearing in the corresponding angular segment around the robot, and $1$ if no robot in in range in that angular segment. The outputs of the neural network are the velocities of the left wheel and the right wheel of the robot, respectively. All sensory activations input to the neural network are scaled in the range $[-1,1]$, while the output neuron activations in range $[-1,1]$ are mapped into wheel velocities in the range $[-10,10] ~\SI{}{cm/s}$. During the evolutionary process, the topology of the neural network as well as the weights of individual connections are evolved, and recurrent connections are allowed in the network (see Table~\ref{tab:evolution-parameters} for details).

\subsection{Tasks for the robot swarm}
Robot swarm controllers are evolved in separate and independent experiments for aggregation, dispersion, flocking, patrolling, and border-patrolling tasks. Multiple evaluation trials are performed for each of the tasks to provide an accurate assessment of fitness. 
\begin{itemize}
\item \textbf{Aggregation:} Robots of the swarm are tasked to form a coherent and stable cluster. The fitness function penalises robots for having a large normalised distance to the centre of mass of the swarm, and averages the penalty over the number of robots in the swarm and the number of control cycles in the evaluation trial:
\begin{equation}
  f_{\text{aggr}} = \frac{1}{T N_r}\sum_{t=1}^{T} \sum_{r=1}^{N_r} 1 - \frac{||{\mathbf{X}_r(t)} - {\mathbf{X}_{cm}(t)}||_{2}}{M} \,,
\end{equation}
where $T$ is the number of control cycles in the evaluation trial, $N_r$ is the number of robots in the swarm, $\mathbf{X}_r(t)$ and $\mathbf{X}_{cm}(t) = \frac{1}{N_r} \sum_{r=1}^{N_r} \mathbf{X}_r(t)$ are the positions of the robot $r$ and the centre of mass of the swarm at control cycle $t$, respectively. Robot distances are normalised by $M$, the length of diagonal of the arena of the swarm.

\item \textbf{Dispersion:} The robots of the swarm are tasked with maximizing their total sensing coverage of the arena,  relevant for the coverage of large areas in patrolling and monitoring missions \cite{Duarte2016,chamanbaz2017swarm}. The fitness function is defined as the normalised distance of all the robots of the swarm to their nearest neighbours, averaged over the number of robots in the swarm and the number of control cycles in the evaluation trial:
\begin{equation}
f_{\text{disp}} = \frac{1}{T N_r}\sum_{t=1}^{T} \sum_{r=1}^{N_r} \frac{||\mathbf{X}_r(t) - \mathbf{X}_{r'}(t)||_{2}}{M/2} \,, 
\end{equation}
where $\mathbf{X}_{r'}(t)$ is the position of the nearest robot to robot $r$ at control cycle $t$. To ensure a tightly bounded fitness in $[0,1]$, normalisation is based on half of the length of the diagonal of the arena of the swarm.

\item \textbf{Flocking:} The swarm is tasked to move in a tightly-coordinated flocking manner. The fitness functions rewards pairs of robots in the swarm within a range of $\SI{50}{cm}$ (half the range of the range-and-bearing sensors), for moving rapidly in the same direction. If at some control cycle $t$, two robots $i$ and $j$ are within range of each other and the angular difference in headings between them satisfies $\Delta \theta_{ij}(t) < 90^{\circ}$, then the robot pair $(i,j)$ is rewarded by $V_i(t) \mbox{ } V_j(t) (1.0 - \Delta \theta_{ij}(t) / 90^{\circ})$,
  where $V_i(t)$ is the linear velocity of robot $i$ in its current orientation, expressed in $[-1,1]$ as a proportion of its maximal linear speed. Otherwise, if $\Delta \theta_{ij}(t) \geq  90^{\circ}$ or $i$ and $j$ are not within range, then a robot pair $(i,j)$ is not rewarded. 
The fitness function considers the average of this reward over the number of control cycles in the evaluation trial:
\begin{equation}
f_{\text{flock}} = \frac{1}{T N_r(N_r - 1)/2}\sum_{t=1}^{T} \sum_{i=1}^{N_r} \sum_{\substack{j > i  \\ j \in R_i(t)} } \big( 1 - \min(1,  \Delta \theta_{ij}(t) / 90^{\circ}) \big) \max(0, V_i(t) \mbox{ } V_j(t))  \,,
\end{equation}
where $\min$ and $\max$ functions return the minimum and maximum, respectively, of their two arguments, and $R_i(t)=\{j \in \{1,\dots,N_r\} \, | \, ||\mathbf{X}_i(t) - \mathbf{X}_{j}(t)||_{2} < \SI{50}{cm} \}$ denotes the robots in range of robot $i$ at control cycle $t$.

\item \textbf{Patrolling:} Robots of the swarm are tasked to actively patrol an arena, a behaviour relevant for surveillance type missions. For the fitness of the patrolling task, the arena of the swarm is discretised uniformly into a grid of cells, initially all given the minimal value of $0$. If one or more robots visit a cell in a control cycle, the value of that cell is set to $1$. Otherwise the cells' values decay continually at a rate $\SI{0.005}{/s}$. The fitness of the patrolling swarm is the average value of all the cells in the arena across the number of control cycles in the evaluation trial:
\begin{equation}
f_{\text{patrol}} = \frac{1}{|C| T}\sum_{t=1}^{T} \sum_{c \in C} c(t) \,,
\end{equation}
where $C$ is the set of cells in the arena and $c(t)$ is the value of the cell at control cycle $t$. In our implementation, the arena is discretised into $10 \times 10$ cells, and $|C|=100$ cells.

\item \textbf{Border-patrolling:} The task, while similar to patrolling, requires the robots of the swarm to patrol along the walls of the arena. For the fitness, the average value of the outermost cells in the arena is considered, across the number of control cycles in the evaluation trial:
\begin{equation}
f_{\text{border-patrol}} = \frac{1}{|C_b| T}\sum_{t=1}^{T} \sum_{c \in C_b} c(t) \,,
\end{equation}
where $C_b \subset C$ is the set of the outermost cells of the arena. For our $10 \times 10$ cells arena $|C_b|=36$ cells.
\end{itemize}

\subsection{Baseline algorithms}
\label{sec: baseline}

Our study compares the proposed QED environment-descriptor algorithm with three baseline algorithms employing task-specific and generic behavioural descriptors. The baseline algorithms are variants of MAP-Elites, employing the following behavioural descriptors: i) a 3-dimensional hand-coded descriptor designed with detailed knowledge of the swarm robotic tasks; ii) SDBC, a 10-dimensional behavioural descriptor, which requires some domain specific knowledge on the task domain, but is otherwise task-generic \cite{Gomes2018a}; and iii) SPIRIT, a completely generic 1024-dimensional behavioural descriptor of the state-action space of the robot \cite{Meyerson2016}. Importantly, as maps generated with the QED algorithm may contain a maximum of $4096$ solutions (six environment attributes and four perturbations per attribute, see Table~\ref{tab:attributeVec}), for comparison, the behaviour-performance maps of the baseline algorithms are discretised to contain the same number of $4096$ solutions.

\noindent\textbf{Hand-coded behavioural descriptor (HBD):} The behavioural descriptor tracks the positions of the robots in a swarm during the evaluation trial, discretised over a uniform grid of cells of size equal to the Thymio robot, to compute the following features: i) the uniformity of the visitation probabilities of different cells in the arena; ii) the average distance of the robots to the center of the arena; and iii) the total number of cells visited by the robots in in the arena at least once during a trial. The obtained features were scaled to range $[0,1]$, discretised into $16$ bins, and averaged across all the evaluation trials to obtain the behavioural descriptor for the swarm.


\noindent\textbf{Systematically Derived Behaviour Characterisations (SDBC):} 
We replicate the 10-dimensional characterisation used in \cite{Gomes2018a}, the mean and standard-deviation over the evaluation trial of the following five features recorded at each control cycle: i) the average linear velocity of the robots in the swarm; ii) the average angular velocity of the robots in the swarm; iii) the average distance between the robots of the swarm and the walls of the arena; iv) the average distance between each and every robot in the swarm; and v) the average distance between each and every robot in the swarm and its closest neighbouring robot. The resulting behavioural descriptor for the swarm is obtained by computing the geometric median of the features across all the evaluation trials. Due to the higher dimensional behavioural descriptor of the SDBC, we employ the Centroidal Voronoi Tesselations (CVT) MAP-Elites algorithm to partition the behaviour space into $4096$ tessellations. Centroids for the CVT MAP-Elites are generated with $100,000$ seed points sampled uniformly in the space $[0,1]^{10}$  (for details see \cite{Vassiliades2018d}). 

\noindent\textbf{Stochastic Policy Induction for Relating Inter-task Trajectories (SPIRIT):} The behavioural descriptor counts the frequencies of sensory states $s \in \mathcal{S}$ and actions $a \in \mathcal{A}$ over all the robots of the swarm, and all the evaluation trials during the simulation, to estimate conditional probabilities $p(a|s)$ for all $a \in \mathcal{A}$ and all $s \in \mathcal{S}$. If a state $s$ has been visited $N$ times, where $N > 0$, then $p(a|s)$ is estimated as the frequency of the state-action pair $(s,a)$ divided by $N$. If a state $s \in \mathcal{S}$ has not been visited, the equiprobable distribution is used, i.e., $p(a|s) = 1/|\mathcal{A}|$ for all $a \in \mathcal{A}$. The action space $\mathcal{A}$ comprises the left and right wheel velocities of the robot, each binned into four equal sized intervals in range $\pm \SI{10}{cm/s}$. The sensor space $\mathcal{S}$ comprises 6 sensor groups, the front-left, front-centre, front-right and rear proximity sensors of the robot, as well as the front and rear facing range-and-bearing sensors, each binarised such that a sensor group is considered active (set to 1) if any one of its sensors' readings exceeds half of the maximum range of that sensor. Therefore, applying SPIRIT to this set-up results in 64 probability distributions of 16 actions each, for a 1024-dimensional characterisation. Due to the high dimensionality of the search space, similar to SDBC, we employ the CVT MAP-Elites algorithm with $4096$ centroids generated from 1 million seed points sampled uniformly from the space $[0,1]^{1024}$,  using an iterative procedure which ensures all probability distributions sum to $1$.

\subsection{Metrics for analysis}
\label{sec: metrics} 
\paragraph{Behavioural diversity:}
To analyse the maps evolved by the quality-diversity algorithms after evolution, behavioural diversity metrics are required. To allow calculating behavioural diversity metrics which are comparable for all algorithms, the solutions of each map are projected to the 1024-dimensional SPIRIT space as a common generic behavioural space. This projection procedure replays the individuals in the normal operating environment, and then records their behavioural descriptor. For calculating behavioural distances in this space, a statistical distance metric is required. With SPIRIT consisting of $|\mathcal{S}|=64$ conditional probability distributions, one for each sensory state $s \in \mathcal{S}$, two probability distributions $p_1$ and $p_2$, corresponding to a given pair of behavioural descriptors in the SPIRIT space, are compared by computing the average of the total variation distances between their conditional probability distributions: 
\begin{equation}
d(p_1,p_2) = \frac{1}{2 |\mathcal{S}|} \sum_{s \in \mathcal{S}} \sum_{a \in \mathcal{A}} | p_1(a|s) - p_2(a|s) | \,.
\end{equation}
\paragraph{Statistical analysis:}
To assess statistical significance, without the need for normality or other parametric assumptions, we compute two non-parametric statistics in pair-wise manner: i) $p$-values from the Wilcoxon rank-sum test, which are based on the summed rank across the different faulty environments, here used to estimate the consistency of the higher ranking of QED compared to a given baseline algorithm; and ii) Cliff's delta \cite{Cliff}, an effect size metric, used here to estimate the extent to which QED's distribution dominates the distribution of a given baseline algorithm, where a value of $1$ indicates complete dominance while a value of $0$ indicates perfectly overlapping distributions and the sign indicates the direction of the effect (positive if QED outperforms the other algorithm; negative otherwise). To categorise the magnitude of Cliff's delta, we make use of the guidelines provided by Vargha \& Delaney (2000) \cite{Vargha2000}, where $|\delta| \geq 0.43$ is considered a large effect, while $|\delta| \in [0.28,0.43)$ and $|\delta| \in [0.11,0.28)$ are considered medium and small effects, respectively. 
\subsection{Experiments}
\label{sec: experiments}
The QED algorithm is compared against the HBD, SDBC and SPIRIT baseline behaviour-descriptors, across each of the five swarm robotic tasks. All our algorithms are implemented in the \textit{sferes2} evolutionary computation framework \cite{MouretJ.-B.andDoncieux2010}. The code for all our experiments, including the QED, HBD, SDBC and SPIRIT quality-diversity algorithms, and the aggregation, dispersion, flocking, patrolling and border-patrolling swarm robotic tasks, is available online\footnote{\url{https://github.com/resilient-swarms/argos-sferes}}.  

Evolutionary experiments are repeated in five separate and independent replicates. Therefore, in total, $4~\mbox{quality-diversity algorithms} \times 5~\mbox{swarm-robotic tasks} \times 5~\mbox{replicates} = 100$ evolutionary experiments are performed. During evolution, each swarm robot controller is evaluated in $50$ independent trials, each of \SI{400}{s} duration. Robots of the swarm and obstacles in the environment, if any\footnote{Obstacles are only included in a subset of the environments generated by QED, and are not part of the normal operating environment.}, are positioned randomly at the start of each trial, and the performance awarded to the evaluated robot controller is the average fitness across all $50$ trials.

\section{Results}
The maps generated by the QED, HBD, SDBC and SPIRIT quality-diversity algorithms are assessed in two phases. In the first phase, the maps are evaluated on the best performance in the normal operating environment, and on the number of unique solutions in the map (see Section~\ref{sec: mapquality}). In the second phase, we introduce sensor/actuator robot faults and environmental perturbations to the swarm and assess, (i) their impact on the performance of the swarm, (ii) how well the swarm is able to recover from them, and (iii) as a metric for \textit{useful diversity}, how behaviourally diverse are the recovery solutions (see Section~\ref{sec: recovery}).

\subsection{Map quality analysis}
\label{sec: mapquality}
Since the aim of quality-diversity algorithms is to evolve a diverse set of high-performing solutions, a large number of function evaluations is required. To meet this demand, while respecting the available computation budget, the quality-diversity algorithms are evolved over 30,000 generations\footnote{A single replicate required about 300-700\SI{}{h} and 900\SI{}{h} of computational time on a 40-cores Intel Xeon Gold 6138 at 2\SI{}{GHz}, for baseline and QED algorithm, respectively.}. This ensured at least a weak convergence in best and average performance for all algorithms on all tasks (see Fig.~S1 in Supplementary Materials).

\paragraph{Performance:} To analyse the quality of individual solutions in the evolved maps, we calculate their performance, defined as the fitness averaged across all independent trials in which it is evaluated. To assess the quality of a map, the best and average performance of a map are then defined by the maximal and mean performance of all its solutions, respectively.  During evolution, the baseline algorithms (HBD, SDBC, SPIRIT) do not vary widely in the best performance. While QED has the highest best performance and an average performance similar to that of baseline algorithms (see Fig. S1 in Supplementary Materials), cautious interpretation is required because QED's perturbed environments have varying levels of difficulty. 

\begin{figure}[htbp!]
\centering
\subfloat[Coverage time-line]{
\includegraphics[width=0.4\textwidth]{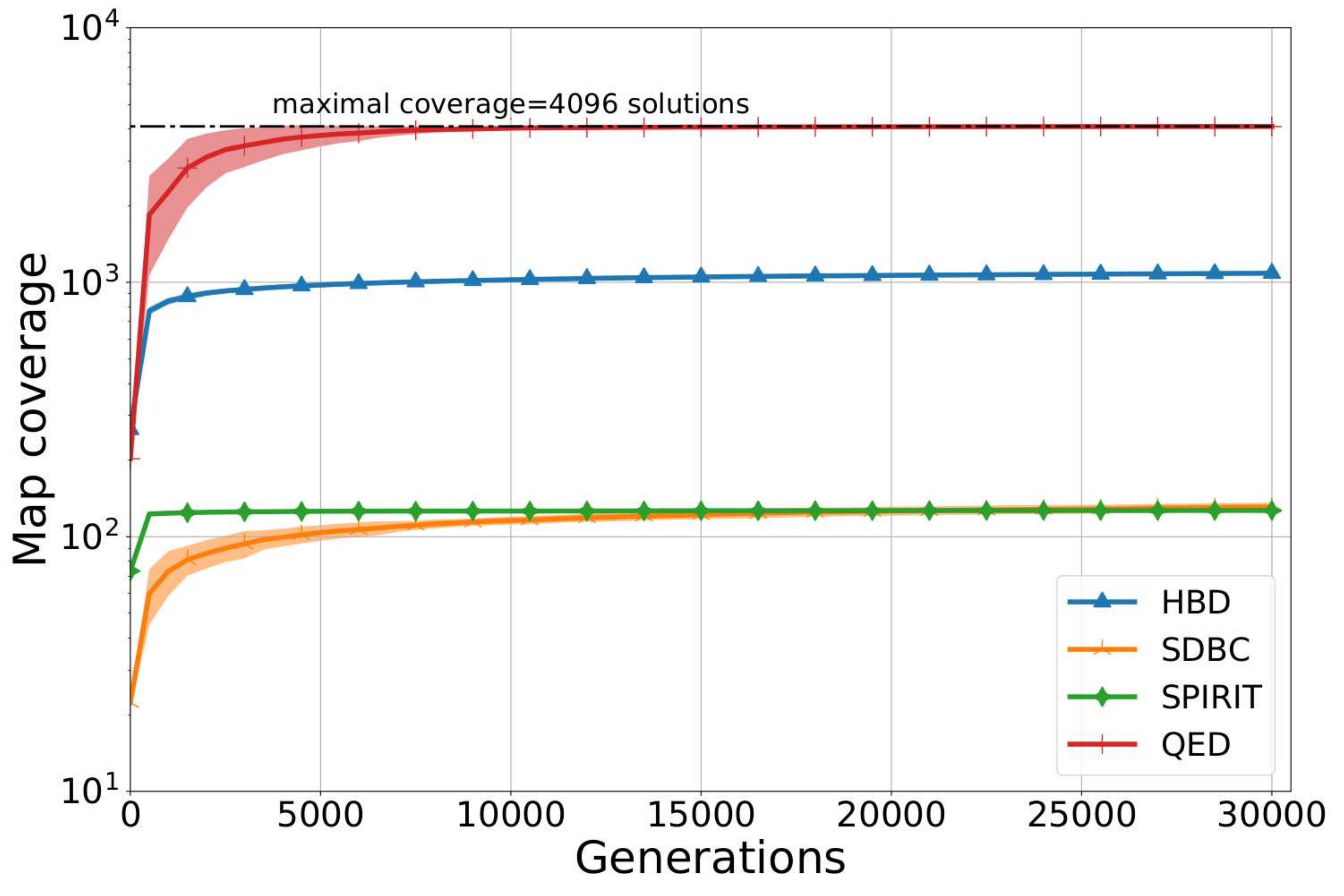} \label{fig: development}}
\subfloat[Performance at end of evolution]{
\includegraphics[width=0.3\textwidth]{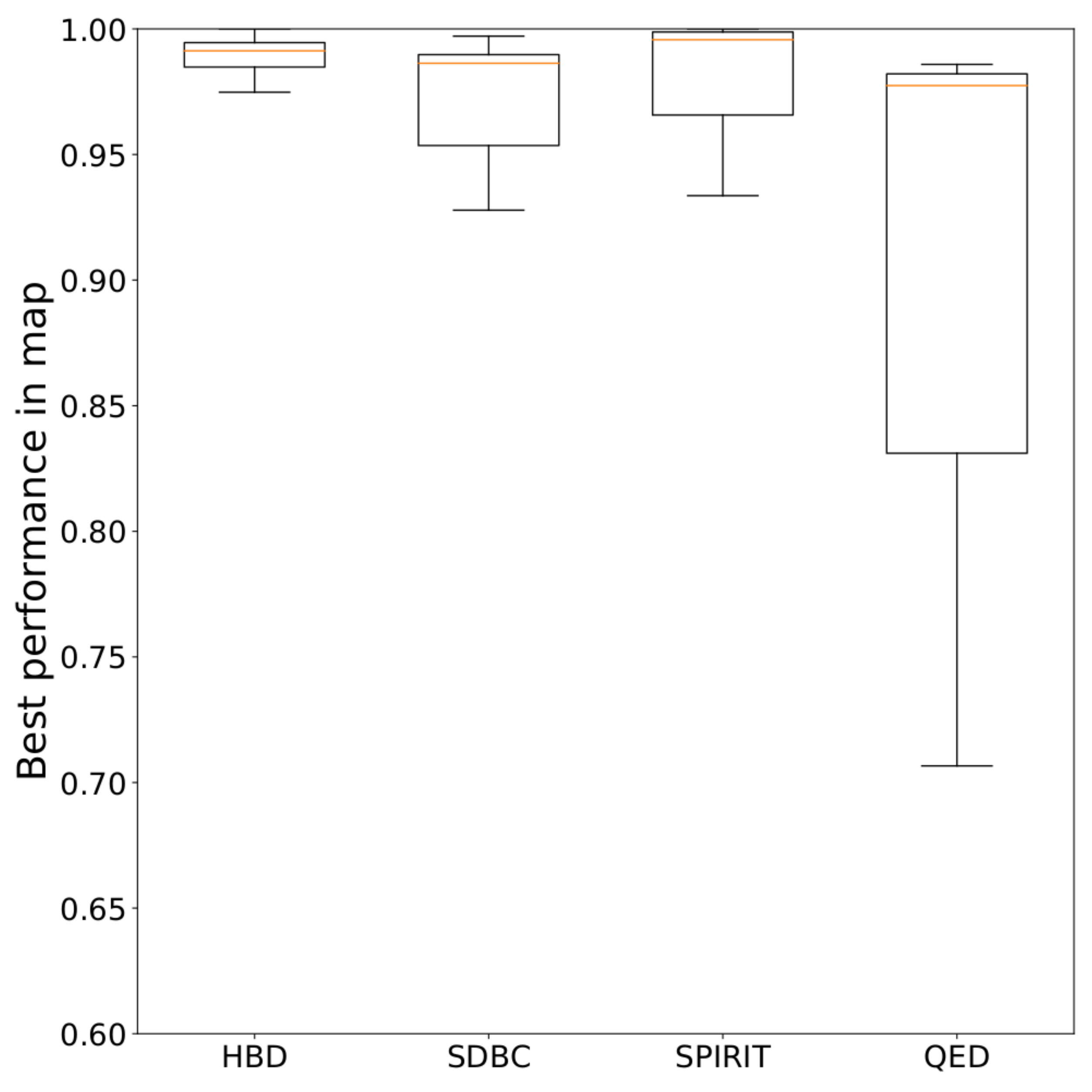} \label{fig: boxplot}
}
\caption{Map quality analysis based on behavioural diversity and performance: (a) development of the map coverage (Mean $\pm$ SD) over generations, with solid line indicating the average across all 25 independently evolved maps (5 independent replicates for each of 5 tasks), and shaded region indicating the standard-deviation across replicates (averaged across all 5 tasks). (b) boxplot of the best performance, evaluated in the normal operating environment; to obtain the best performance data for a given map, all its individual solutions are re-evaluated for 10 independent trials in the normal operating environment; for any given algorithm, each data point in the box plot represents one of the 25 re-evaluated maps\protect\footnotemark.} \label{fig: after-evol-statistics}
\end{figure}

To allow a fair performance comparison between algorithms, we must estimate the best and average performance of QED maps \textit{after} evolution, by evaluating their solutions in the normal operating environment\footnote{Note that, in QED, the best solution to the normal operating environment may be hidden anywhere in the environment-performance map; therefore, to obtain the best performance it is not sufficient to simply take the elite solution for the normal operating environment in QED's environment-performance map.}. Due to the large number of solutions in QED maps, 50 trials of fitness evaluation would exceed our computational budget. Therefore, the performances of solutions evolved by QED are all re-evaluated based on 10 trials each and, to keep the performance calculations comparable, the solutions evolved by the other algorithms (HBD, SDBC, and SPIRIT) are also re-evaluated based on 10 trials. This analysis demonstrates (see Table S1 in the Supplementary Materials) that for aggregation, patrolling and border-patrolling, all the algorithms are close in best performance, differing only by at most 3\% of the empirical maximum performance\footnote{The empirical maximum performance of a task is the maximal performance observed across all replicates and all algorithms.}. However, the performance varies more widely in flocking and dispersion: in the flocking task, the best performance ranges widely (25-85\% for QED, 70-98\% for SDBC, and 66-100\% for SPIRIT), with the exception of HBD which consistently performs the empirical maximum (97-99\%); in the dispersion task, QED has a best performance of 81-87\% while the baseline algorithms (HBD, SDBC, and SPIRIT) have a best performance between 93-100\%. Joining the data from all tasks together (see Fig. \ref{fig: boxplot}),  we find that QED sacrifices best performance in the normal operating environment for improved best performance on the environments in which it evolved; in other words, the other algorithms may overfit on the normal operating environment. In this sense, the HBD is particularly specialised towards maximising performance in the normal operating environment.

As a score for the overall performance of solutions in the evolved maps, we compute for each algorithm the average performance of their maps and then take the mean across the independent replicates. We find that QED has a low overall performance for dispersion, with a score of 60\%  compared to a score ranging in 70-90\% for baseline-algorithms,  and flocking, with a score of 24\% compared to a score ranging in 20-40\% for other algorithms; however, QED is second in rank on the other tasks, where its score is around 90\%. SPIRIT obtains the highest overall performance on all tasks, with scores all greater than 90\%, with the exception of its low score of 44\% on the flocking task. These findings indicate that SPIRIT, and, to a lesser degree, QED, can find a variety of controllers with high performance on the normal operating environment. 

\paragraph{Behavioural diversity:} As an approximate indicator of behavioural diversity in the algorithms' own behaviour spaces, we compute the coverage (see Fig.~\ref{fig: development}), the number of cells in an algorithm's own behaviour space that are filled with a solution. At the end of evolution, the coverage (Mean $\pm$ SD) is $4094 \pm 2$ for QED, $1091 \pm 19$ for HBD, $130 \pm 5$ for SDBC, and $127 \pm 1$ for SPIRIT. These results indicate that the behaviour spaces of HBD, SPIRIT and SDBC are sparse, meaning that not all cells represent feasible robot swarm behaviours, whereas the behaviour space of QED is dense, with all cells representing feasible robot swarm behaviours. These findings are not sufficient, however, to demonstrate that QED is behaviourally diverse, since the QED space is based on \textit{environmental diversity}.

To allow a comparison of behavioural diversity, we compute behavioural diversity metrics based on a common behaviour space. To do so, we first project the solutions from the different maps obtained by evolution to a common generic behaviour space, as explained in Section \ref{sec: metrics}. Then, since the resulting projected maps may have more than one solution for some of the centroids, these projected maps are further processed into a valid behaviour-performance map by maintaining at most one elite solution for each of the 4096 centroids. The behaviour of each elite solution is then described using the corresponding centroids. Computing the average pair-wise distance between the non-empty cells in the projected behaviour-performance maps as the behavioural diversity score, it is observed that all algorithms have a similar behavioural diversity ranging between 0.60-0.62; however, the coverage (Mean $\pm$ SD) in the projected behaviour-performance map differs more widely, with SDBC ($26 \pm 2$) scoring just half of HBD ($55 \pm 10$), SPIRIT ($51 \pm 11$), and QED ($49 \pm 20$). Based on these findings, the QED space, which is based on an environment descriptor, yields a behavioural diversity at least on par with the behaviour spaces of the baseline-algorithms (HBD, SDBC, and SPIRIT), which are based on behavioural descriptors.


\footnotetext{The middle line in the box indicates the median, while the box ranges from the first to the third quartile of the data. Whiskers are defined by the highest data-point still within 1.5 times the IQR of the third quartile and the lowest data-point still within 1.5 times the IQR of the first quartile.}

\subsection{Fault recovery analysis}
\label{sec: recovery}
A traditional analysis of behavioural diversity is limited in the sense that behavioural diversity in itself is not a sufficient condition for fault recovery or adaptation; there may be behaviours which contribute to diversity but not to fault recovery. Similarly, a high performance in the normal operating environment does not necessarily generalise towards environments not seen during evolution. Therefore, we analyse behavioural diversity and performance within the context of robot swarm fault recovery, where all robots in the swarm may have different faults. 
\paragraph{Fault injections:}
When a robot swarm is operating in its normal operating environment, any fault to a single robot's sensors or actuators effectively changes the operating environment. Because different robots in the same swarm may have different faults, the space of possible environments is greatly expanded. Here we explore a limited region of this vast space by randomly injecting independent faults to each of the robots in the swarm. For each robot in the swarm $i \in \{1,\dots,10\}$, we select randomly one out of 8 faults, leading to a combined fault $\mathscr{F}$ and the corresponding environment $\mathcal{E}_{\mathscr{F}}$ in which the performance is then assessed. 

The fault injection scheme uses the following faults, as used in previous studies on fault-detection in robot swarms  \cite{Tarapore2017,Tarapore2019}:
\begin{itemize}
\item \textbf{PMIN}: the front proximity sensor readings are set to the minimum (0) at each control cycle;
\item \textbf{PMAX}: the front proximity sensor readings are set to the maximum (1) at each control cycle;
\item \textbf{PRAND}: the front proximity sensor readings are generated randomly in the range $[0,1]$, at each control cycle;
\item \textbf{LW-H}: the speed of the robot's left wheel is halved, at each control cycle;
\item \textbf{RW-H}: the speed of the robot's right wheel is halved, at each control cycle;
\item \textbf{BW-H}: the speed of both wheels is halved, at each control cycle;
\item \textbf{ROFS}: a large offset vector $(r,\theta)$ is added to range-and-bearing readings, with $r \sim U(0.75,1.0)$ and $\theta \sim U(-180^{\circ},180^{\circ})$, at each control cycle;
\item \textbf{NONE}: no fault is applied.
\end{itemize}
Importantly, none of the resulting environments are seen during the evolutionary phase by any of the algorithms, and the fact that different robots experience different faults is particularly challenging.

In our experiments, a total of 250 unique combined faults are sampled and applied to all 5 tasks; this means there are 1250 samples to assess diversity and fault recovery for each quality-diversity algorithm. Based on these samples, the faults are distributed across the swarm as follows: range-and-bearing sensor faults affect 0-4 robots; proximity and range-and-bearing sensor faults each affect 1-7 robots; and, 7-10 robots are affected by faults other than NONE. The 250 combined faults are divided evenly amongst 5 independent maps. Due to the large number of faulty environments and the large number of controllers in the maps, the number of trials for the fitness evaluation is set to 10. 

%

\paragraph{Fault recovery:}  
In our model-free approach, fault recovery is to be achieved by searching the map for the best solution to the faulty environment. 
To assess map quality in the context of fault recovery, we estimate performance, comparing the performance in the faulty and normal operating environments, as well as useful diversity, the behavioural diversity across the recovery solutions.

The analysis of the fault recovery includes a total of three metrics: i) the impact, the proportional change in performance after transferring the best solution of the normal operating environment to the faulty 
environment;  ii) the recovered performance,  the performance of the best 
solution to the faulty environment; and iii) the resilience $\mathscr{R}$, the proportional change in performance comparing the best solution for  the faulty environment $\mathcal{E}_{\mathscr{F}}$ to the best solution for the normal operating environment $\mathcal{E}$:  
\begin{equation}
\label{eq: resilience}
\mathscr{R}(\mathcal{M},\mathcal{E}_{\mathscr{F}} | \mathcal{E}) = \frac{ \max_{c \in \mathcal{M}} f(\mathcal{E}_{\mathscr{F}},c) - \max_{c' \in \mathcal{M}} f(\mathcal{E},c')}{\max_{c' \in \mathcal{M}} f(\mathcal{E},c')} \,,
\end{equation}
where $\mathcal{M}$ is the map.

A summary of these metrics, using the median and inter-quartile range across all fault injections, demonstrates the viability of our approach to fault recovery (see Table S2 in Supplementary Materials). All quality-diversity algorithms have a median recovered performance between 86-88\% of the empirical maximum performance.  Despite the median impact of the fault being 23-25\% of the original performance in the normal operating environment, the median resilience score indicates that after fault recovery, the drop in performance is only 8-12\%. These results also demonstrate differences between the algorithms: QED has the highest score (Median $\pm$ IQR) for impact of the fault ($-0.23 \pm 0.26$), recovered performance ($0.88 \pm 0.16$),  and resilience ($-0.08 \pm 0.07$), whereas SPIRIT has the lowest score for impact of the fault ($-0.25 \pm 0.29$), recovered performance ($0.86 \pm 0.13$), and resilience ($-0.12 \pm 0.13$). Explaining its high IQR, the impact of the fault depends strongly on the type of task, ranging between 0-20\% for aggregation, 10-40\% for dispersion, 10-50\% for patrolling, 0-60\% for border-patrolling, and 60-100\% for flocking. After plotting the number of robots affected by faults to proximity sensors, range-and-bearing sensors, or actuators, no direct relation is observed between fault type and the impact of the fault.

Since the resilience indicates to what extent the swarm's performance will be affected after recovery, and the recovered performance indicates the quality of the recovery solution, these two scores are arguably of most interest to the end-user. For each task, we perform pair-wise comparisons of resilience and recovered performance, comparing QED to the baseline-algorithms, using the statistical analysis mentioned in Section \ref{sec: metrics}, based on $n=250$ data points per algorithm (see Table S3 in Supplementary Materials). On the resilience score, QED significantly outperforms all other algorithms on all tasks with medium to large effect sizes, with only two exceptions, namely HBD in aggregation ($p=0.396$, $\delta=0.04$) and SDBC in patrolling ($p=0.038$; $\delta=0.17$). The recovered performance results are more mixed, and QED outperforms other algorithms 9 out of 15 times, but with varying significance and effect sizes. Patrolling and border-patrolling are solved best by QED, with medium to large effect sizes. Similar to the results in the normal operating environment, QED has a low performance on the dispersion task, with a large negative effect in this case. Other observations from the task-specific analysis in Table S2 are that: i) HBD's median performance varies, with the highest scores across algorithms on aggregation, dispersion, and flocking, but the lowest and second lowest on border-patrolling and patrolling, respectively; ii) the flocking task is not satisfactorily solved by any algorithm, with median recovered performance ranging in 17-24\% of the empirical maximum performance. 

\begin{figure*}[htbp!]
\centering
\subfloat[Impact-resilience signature]
{
\includegraphics[width=1.0\textwidth]{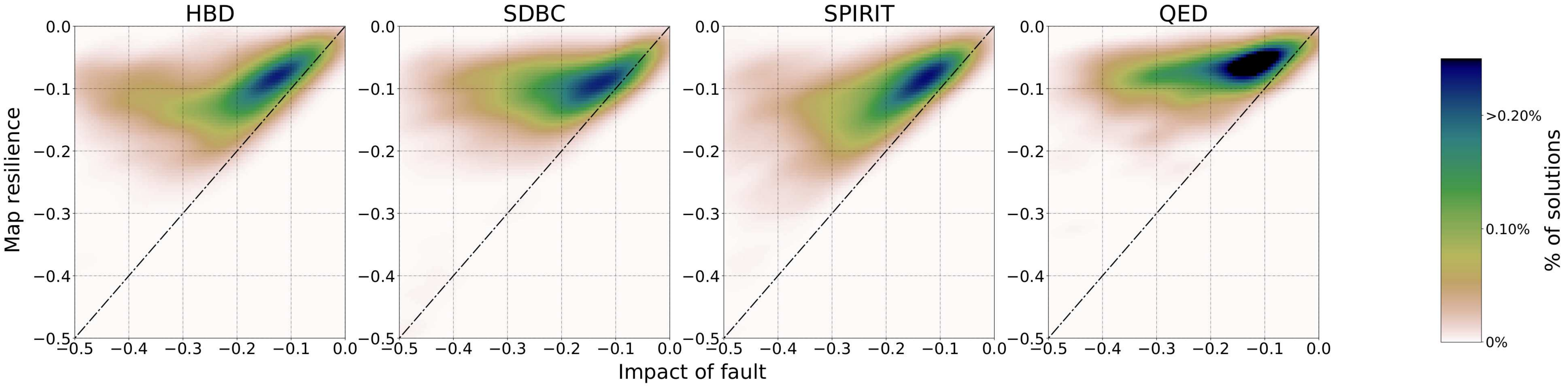} \label{fig: impact-resilience}
}
\\
\subfloat[Diversity-resilience signature]
{
\includegraphics[width=1.0\textwidth]{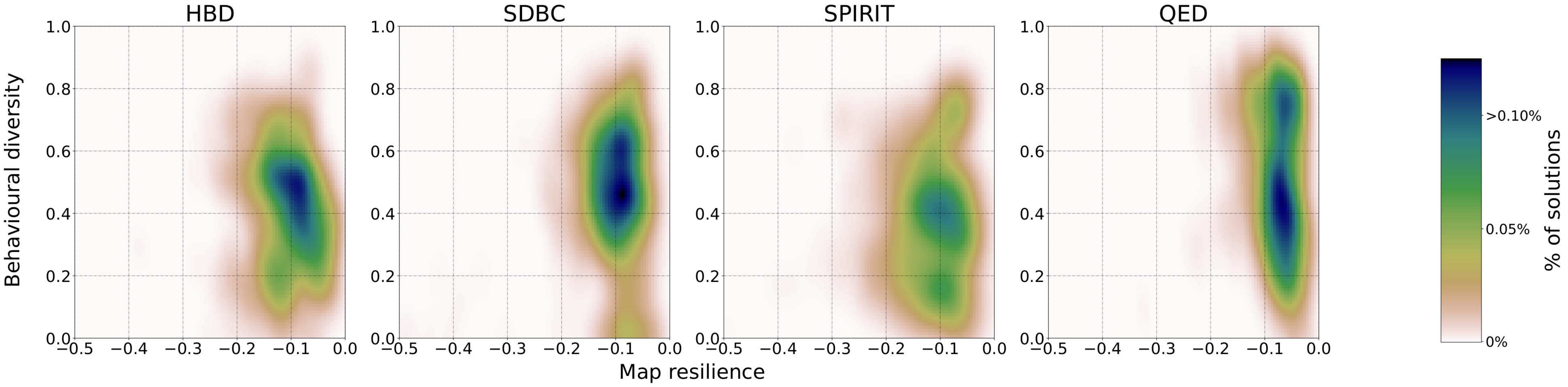} \label{fig: diversity-resilience}
}
\\
\subfloat[Diversity-impact signature]
{
\includegraphics[width=1.0\textwidth]{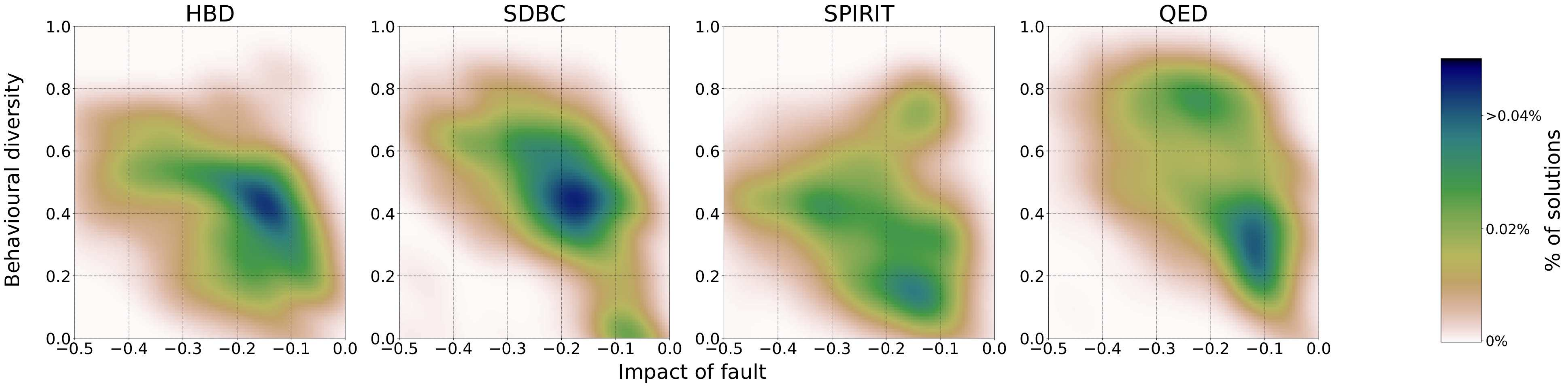} \label{fig: diversity-impact}
}
\caption{Visualisation of the fault recovery data based on impact of the fault, resilience, and behavioural diversity around the normal behaviour: (a) impact-resilience signature, illustrating the probability distribution of resilience as a function of the impact of the fault, with the dashed line indicating the identity line. An ideal signature has most of its probability mass in the top right corner, where faults do not affect the swarm's performance; however, in case there are high-impact faults, ideally the resilience should be constant across the impact spectrum, indicating that recovery will be possible even for high-impact faults. (b) diversity-resilience signature, illustrating the probability distribution of behavioural diversity, based on the behavioural distance to the normal behaviour, as a function of resilience. An ideal signature has high resilience and covers the entire distance spectrum, indicating a strong fault recovery profile with useful diversity. (c) diversity-impact signature, illustrating the probability distribution of behavioural diversity, based on the behavioural distance to the normal behaviour, as a function of the impact of the fault. Each of the signatures illustrates a correlation between the impact of the fault and the behavioural diversity, indicating behaviours that differ strongly from the normal behaviour allow improved recovery from high-impact faults\protect\footnotemark.} \label{fig: signature}
\end{figure*}

The resilience score is expected to be become increasingly important for high-impact faults, because algorithms which cannot recover from high-impact faults will have their performance severely affected. Therefore, to predict an algorithm's fault recovery performance in the face of unknown faults, we visualise its resilience as a function of the impact of the fault, giving each algorithm its unique \textit{impact-resilience signature} (see Fig. \ref{fig: impact-resilience})\footnote{To better visualise the majority of the data at a fine resolution and allow meaningful estimates of the slope which are not affected by extreme observations, the observations with extreme impacts (i.e., those with impacts larger than 50\%) are removed from the analysis of Fig. \ref{fig: signature}, resulting in around 1000 data points for each algorithm.}. This analysis demonstrates that QED's comparative resilience advantage derives not only from having lower impacts, typically giving only a 10\% drop in performance, but also from being more resilient to high-impact faults; although all algorithms have a positive correlation between impact and resilience (0.4-0.6), QED has a smaller slope ($a=0.17$) compared to other algorithms ($a=0.19$ for HBD, $a=0.19$ for SDBC, and $a=.30$ for SPIRIT) when considering a linear regression model of resilience as a function of impact.

To explain its improved resilience to high-impact faults, our hypothesis is that QED overcomes high-impact faults by finding high-performing fault recovery behaviours that are especially different from the \textit{normal behaviour}, i.e., the behavioural descriptor obtained from the best-performing solution to the normal operating environment. To assess this interpretation, the resilience and the behavioural diversity of the fault recovery solutions are visualised in Fig. \ref{fig: diversity-resilience}. Confirming its resilience property, QED has typical values around $-0.07$ with most of its probability mass in $[-0.10,0]$  while other algorithms have a resilience centred around $-0.10$ with most of the probability mass in $[-0.20,0]$. Importantly, QED exhausts nearly the entire spectrum of distances around the normal behaviour, while other algorithms have a maximal distance of 0.7-0.8. As a signature that is most opposite to QED, SPIRIT has both low diversity and resilience. Combining the diversity data with the above-mentioned slope data, where resilience degrades slowly for QED but swiftly for SPIRIT as the fault impact grows larger, provides a first supporting argument for our hypothesis. 

To further demonstrate that the solutions that are highly differing from the normal behaviour are frequently used for recovery from high impact faults, we provide a visualisation of behavioural diversity as a function of the impact of the fault. This analysis  (see Fig. \ref{fig: diversity-impact}) confirms that faults with a large, negative impact on performance are more likely to yield solutions with a higher behavioural distance to the normal behaviour. The visualisation is supported by the negative correlations between impact of the fault and behavioural diversity and a large negative slope for all algorithms ( $a=-0.66$ for HBD, $a=-1.01$ for SDBC, $a=-0.36$ for SPIRIT, and $a=-0.98$ for QED) when considering a linear regression model of behavioural diversity as a function of fault. Another observation is that, although baseline algorithms (HBD, SDBC, and SPIRIT) have larger fault impacts, some of which exceed a drop of 40\% of the original performance, they only rarely find solutions at high behavioural distances ($d>0.70$) to the normal behaviour, whereas QED does so frequently for faults with an impact greater than 20\%. These findings confirm our hypothesis because, when faced with high-impact faults, QED more frequently finds solutions that strongly differ from the normal behaviour.
\footnotetext{The proportion estimates are based on a Gaussian kernel density estimation with 100-by-100 grid size and bandwidth according to Scott's rule, $n^{-1/(d+4)}$, where $n$ is the number of data points and $d$ is the dimensionality, here $n=1250$ and $d=2$.}

To provide a visual demonstration of the results, the above-mentioned relation between high-impact faults, resilience, and behavioural diversity around the normal behaviour is further supported by video footage of the border-patrolling task\footnote{See \url{https://youtu.be/BN6i-NugCGg} for an example fault recovery behaviour for the swarm performing a border-patrolling task.}. In this footage, all algorithms  (HBD, SDBC, SPIRIT, and QED) can reduce the impact of faults to the front proximity sensors by using a controller which drives backwards and relies on the rear proximity sensors. Another common observation for all algorithms is that the robots with one faulty wheel make circular movements at a fixed position. However, in the baseline algorithms (HBD, SDBC, and SPIRIT), such robots interfere with the entire swarm, resulting in a traffic congestion, while in QED these robots are isolated and other robots in the swarm stay close to the walls and avoid collision. We believe this is in a large part due to the environment in which the solution was evolved: with a larger number of robots each having a higher speed, the robots in the swarm came in frequent contact with each other, and therefore avoiding other robots was important to obtain a high performance. This footage is in line with the above-mentioned findings: despite the high impact of the fault, QED drops a mere 2\% in performance, using a recovery solution which has a behavioural distance of 0.8 to its normal behaviour, while other algorithms drop 7-8\% and have a distance between 0.6-0.7 to their normal behaviour.

\section{Discussion}

This paper investigates the use of quality-diversity algorithms in a model-free approach to fault recovery in swarm robotic systems, where different robots in a swarm are affected by different faults. Using algorithms based on MAP-Elites, we provide a comparative study of four different behavioural descriptors across five different swarm robotics tasks. Our novel Quality-Environment-Diversity (QED) framework, which describes the behaviour of individual solutions implicitly based on the environment in which they are evaluated, is compared to different baseline-algorithms, including a domain-specific, hand-coded behavioural descriptor, as well as two generic behavioural descriptors, SDBC and SPIRIT. Overall, the study demonstrates the use of quality-diversity algorithms as a means for fault recovery in swarm robotics; the quality-diversity algorithms are typically able to reduce the loss of performance from 23-25\% to 8-12\% of the performance before the fault injections. The study further demonstrates the resilience property of the environment-performance maps evolved by QED, as the solution after fault recovery is similar in performance to the solution before the faults were injected to the robots in the swarm. Due to evolving solutions in a wide diversity of environments, QED environment-performance maps contain fault recovery solutions which differ strongly from the best solutions in the normal operating environment, allowing improved adaptation to high-impact faults.

The results of our investigation provide us with several recommendations. Based on the large resilience advantage and the favourable diversity features of QED, environment diversity is an important component to be considered when evolving a diversity of solutions to a given optimisation problem; this recommendation likely extends beyond the use case of swarm robotics and fault recovery. However, a trade-off to consider is that performance in the normal operating environment may in some cases be significantly reduced by using QED. In fact, in the dispersion task, QED's resilience property was not sufficient to overcome its low performance in the normal operating environment. Our study also explores a hand-coded behavioural descriptor, HBD, which exploits the prior knowledge of the arena used in all the tasks. Despite its comparatively high fault recovery performance in aggregation, dispersion, and flocking, HBD has a comparatively low fault recovery performance in patrolling and border-patrolling. Since one of HBD's behavioural features is the number of unique cells visited in a fine resolution grid, we suggest that the diversity of high-performing fault recovery solutions is limited by the high correlation of HBD's descriptor to the patrolling and border-patrolling fitness functions. Therefore, for fault recovery in swarm robotics, we cautiously advise against the use of behavioural descriptors that are `aligned' with the fitness function, contrasting to results in deceptive maze problems \cite{Kistemaker2011,Pugh2016}. Further, our study finds that, in the flocking task, fault recovery was not satisfactory for any of the quality-diversity algorithms, with a loss of 60-80\% of the original performance levels. Based on this observation, the recommendation is that, when applying our model-free approach to fault recovery in tightly-coordinated tasks involving coordinated movement of robots \cite{Viragh2014,Gross2009,llenas2018quality}, each robot should use its own unique controller; this would enable fault recovery in case different robots are affected by different faults. For example, when one robot is affected by a fault on its right actuator, and another robot is affected by a fault on its left actuator, these two robots can only move coordinately when they have a different controller. However, using a unique controller for each different controller results in at least two challenges: first, there is a linear increase in the number of dimensions in the genotypic space, and therefore an exponential increase in the search space; second, for descriptors defined based on the policy of a single robot, such as the SPIRIT descriptor, this also results in an exponential increase in the number of behaviours to be considered, unless the resolution of the behaviour-performance map is significantly reduced.


Only limited work has exploited environment diversity to generate behaviourally diverse archives of controllers, with no work so far in swarm robotics or fault recovery; 
QED provides an alternative to existing approaches and is characterised by applicability, scalability, and user-control. One approach which uses environment diversity within quality-diversity algorithms is the Innovation Engines 
\cite{Nguyen2016}, which automatically extracts features from behaviour but scores the fitness dependent on the behaviour characterisation, and in this sense the task objective and the behavioural description are mutually dependent. While Innovation Engines has been applied to generate 
images, this method may not be applicable to evolutionary robotics, where the performance of a robot must be assessed empirically in expensive 
simulations rather than through a neural network classification. Although not explicitly mentioned as a quality-diversity algorithm, the Combinatorial Multi-Objective Evolutionary Algorithm \cite{Huizinga2018} 
stores a multitude of elite solutions for different task-subtask decompositions and evaluates new offspring on each of the decompositions. In comparison, the QED framework is not limited to combinatorial tasks, and, due to the use of the 
environment generator, may be able to apply evolution on a wider diversity of environments. Finally, the Paired Open-Ended Trailblazer (POET) 
\cite{Wang2019} generates an ever-increasing diversity of environments within a certain difficulty level, and pairs each environment with a 
single solution which is optimised on this environment, not based on population-based methods but rather based on local optimisation. With its emphasis on challenging environments, empirical results demonstrate solutions 
to increasingly challenging single-robot control problems. POET may not be applicable to expensive cost functions due to the need to evaluate 
all individuals on all environments (to avoid local minima on existing environments and to initialise the best solution to a new environment); 
in swarm robotics, this may be problematic because high-fidelity simulations are required with many trials, long evaluation times and many 
robots. In comparison to POET, QED provides an increased user-control since environments are generated by a user-defined probability 
distribution, rather than by incremental mutations on existing environments. Further, QED may be preferable when the space of environments must 
be exhausted since POET rejects environments if they lead to either too high or too low fitness scores. In fault recovery, this may imply that 
POET rejects environments where robots are affected by faults with a strong impact on performance.

A point of discussion is to what extent the fault recovery results in simulation environments can be extrapolated to real-world swarm robotic systems.  While the present study assumes that obtaining the best-performing controller in the archive comes at no cost, in real-world swarm robotics tasks it must be assumed that only a limited number of function evaluations. However, as demonstrated by Cully et al. (2015) in a single-robot study, Bayesian Optimisation represents one approach to search efficiently for the best solution in a behaviour-performance map \cite{Cully2015,Shahriari2016}. This approach may potentially work in robot swarms as well, and in this case, QED may provide an alternative to existing behavioural descriptors to improve fault recovery. As a potential alternative to Bayesian Optimisation, with no restrictions on the map's geometry and no need for expensive trial and error, the topologically organised environment-performance map of QED may be exploited by first detecting the environment and then taking the corresponding controller in the map. An important point of interest is that QED may have potential for improved robustness to the simulation-reality gap, because systems which are robust to a transfer from one simulation environment to another may also be robust to a simulation-to-reality transfer \cite{Ligot2019}, and environment diversity has been used previously in the control of a robotic arm to improve the simulation-reality gap \cite{IlgeAkkaya}.

Our fault recovery study also presents a novel method to evaluate the quality of archives developed by quality diversity algorithms. Traditionally, archives evolved by quality-diversity algorithms are evaluated based on behavioural diversity and performance metrics obtained from evolution \cite{Mouret2015,Cully2018}. However, because the designer cannot predict the types of conditions that may arise in real-world application, this analysis is limited. To predict how well an archive will fare in adverse conditions, where faults strongly impact the performance of the swarm, our study has considered a novel model selection tool, called the impact-resilience signature. The signature provides a unique profile for a quality-diversity algorithm by visualising  how strongly performance is expected to degrade in the face of high-impact faults. To further investigate the causes of the differences between algorithms, we similarly visualise for each algorithm its unique profile of diversity as a function of impact and resilience; in our study, this analysis demonstrates that QED can improve resilience by finding behaviours which differ strongly from the normal behaviour.

\section{Conclusion}
We investigate fault recovery in swarm robotics, where each robot in a swarm may be affected by different faults. Our approach is to generate a behaviourally diverse archive of behaviours using quality-diversity algorithms. We formulate a novel quality-diversity algorithm, Quality-Environment-Diversity (QED), which varies environments randomly and which uses an environment-descriptor to characterise behaviour implicitly based on the environment in which it is evaluated. A study including QED and quality-diversity algorithms based on behavioural descriptors demonstrates: i) a successful fault recovery for all quality-diversity algorithms, with the impact on the swarm's performance typically being reduced by a factor 2-3; and ii) the comparatively favourable profile of QED with its high robustness to faults and the ability to generate behaviourally diverse fault-recovery solutions. In future work, the QED framework may be implemented with alternative environment generators or other quality-diversity algorithms such as Novelty Search with Local Competition \cite{LehmanStanley2011}, and fault recovery performance may be targeted more directly by an adaptive selection of environments during the evolutionary process.

\bibliography{library} 

\begin{thebibliography}{10}
\providecommand{\url}[1]{#1}
\csname url@rmstyle\endcsname
\providecommand{\newblock}{\relax}
\providecommand{\bibinfo}[2]{#2}
\providecommand\BIBentrySTDinterwordspacing{\spaceskip=0pt\relax}
\providecommand\BIBentryALTinterwordstretchfactor{4}
\providecommand\BIBentryALTinterwordspacing{\spaceskip=\fontdimen2\font plus
\BIBentryALTinterwordstretchfactor\fontdimen3\font minus
  \fontdimen4\font\relax}
\providecommand\BIBforeignlanguage[2]{{%
\expandafter\ifx\csname l@#1\endcsname\relax
\typeout{** WARNING: IEEEtran.bst: No hyphenation pattern has been}%
\typeout{** loaded for the language `#1'. Using the pattern for}%
\typeout{** the default language instead.}%
\else
\language=\csname l@#1\endcsname
\fi
#2}}

\bibitem{Sahin2005}
E.~Sahin, ``{Swarm robotics: From sources of inspiration to domains of
  application},'' in \emph{Swarm Intelligence. Natural Computing Series.},
  E.~Sahin and W.~Spears, Eds.\hskip 1em plus 0.5em minus 0.4em\relax Springer,
  Berlin, Heidelberg, 2005, vol. 3342, pp. 10--20.

\bibitem{Brambilla2013a}
M.~Brambilla, E.~Ferrante, M.~Birattari, and M.~Dorigo, ``{Swarm robotics: A
  review from the swarm engineering perspective},'' \emph{Swarm Intelligence},
  vol.~7, no.~1, pp. 1--41, 2013.

\bibitem{Bayindir2016}
L.~Bayindir, ``{A review of swarm robotics tasks},'' \emph{Neurocomputing},
  vol. 172, pp. 292--321, 2016.

\bibitem{Schwager2007}
M.~Schwager, J.~McLurkin, and D.~Rus, ``{Distributed coverage control with
  sensory feedback for networked robots},'' \emph{Robotics: Science and
  Systems}, vol.~2, pp. 49--56, 2007.

\bibitem{Duarte2016}
M.~Duarte, V.~Costa, J.~Gomes, T.~Rodrigues, F.~Silva, S.~M. Oliveira, and
  A.~L. Christensen, ``{Evolution of collective behaviors for a real swarm of
  aquatic surface robots},'' \emph{PLoS ONE}, vol.~11, no.~3, pp. 1--25, 2016.

\bibitem{Garattoni2018}
L.~Garattoni and M.~Birattari, ``{Autonomous task sequencing in a robot
  swarm},'' \emph{Science Robotics}, vol.~3, no.~20, pp. 1--12, 2018.

\bibitem{Viragh2014}
C.~Vir{\'{a}}gh, G.~V{\'{a}}s{\'{a}}rhelyi, N.~Tarcai, T.~Sz{\"{o}}r{\'{e}}nyi,
  G.~Somorjai, T.~Nepusz, and T.~Vicsek, ``{Flocking algorithm for autonomous
  flying robots},'' \emph{Bioinspiration and Biomimetics}, vol.~9, no.~2, 2014.

\bibitem{Gross2009}
R.~Gro{\ss} and M.~Dorigo, ``{Towards group transport by swarms of robots},''
  \emph{International Journal of Bio-Inspired Computation}, vol.~1, no. 1-2,
  pp. 1--13, 2009.

\bibitem{llenas2018quality}
A.~F. Llenas, M.~S. Talamali, X.~Xu, J.~A.~R. Marshall, and A.~Reina,
  ``{Quality-Sensitive Foraging by a Robot Swarm Through Virtual Pheromone
  Trails},'' in \emph{International Conference on Swarm Intelligence}.\hskip
  1em plus 0.5em minus 0.4em\relax Springer, 2018, pp. 135--149.

\bibitem{Bonabeau1999}
E.~Bonabeau, M.~Dorigo, and G.~Theraulaz, \emph{{Swarm Intelligence: From
  Natural to Artificial Systems.}}\hskip 1em plus 0.5em minus 0.4em\relax New
  York, NY: Oxford University Press, 1999.

\bibitem{Bjerknes}
J.~D. Bjerknes and A.~F.~T. Winfield, ``{On Fault Tolerance and Scalability of
  Swarm Robotic Systems},'' in \emph{Distributed Autonomous Robotic Systems,
  STAR}, A.~Martinoli, F.~Mondada, N.~Correll, G.~Mermoud, M.~Egerstedt, M.~A.
  Hsieh, L.~E. Parker, and K.~Stoy, Eds.\hskip 1em plus 0.5em minus 0.4em\relax
  Springer-Verlag Berlin Heidelberg, 2013, pp. 431--443.

\bibitem{Winfield2006}
A.~F. Winfield and J.~Nembrini, ``{Safety in numbers: Fault-tolerance in robot
  swarms},'' \emph{International Journal of Modelling, Identification and
  Control}, vol.~1, no.~1, pp. 30--37, 2006.

\bibitem{Tarapore2015}
\BIBentryALTinterwordspacing
D.~Tarapore, P.~U. Lima, J.~Carneiro, and A.~L. Christensen, ``{To err is
  robotic, to tolerate immunological: fault detection in multirobot systems},''
  \emph{Bioinspiration {\&} Biomimetics}, vol.~10, no.~1, p. 016014, 2015.
  [Online]. Available:
  \url{http://stacks.iop.org/1748-3190/10/i=1/a=016014?key=crossref.6b4f84626c1e0899d7c26b606855ffe1}
\BIBentrySTDinterwordspacing

\bibitem{Christensen2008}
A.~L. Christensen, R.~O'Grady, M.~Birattari, and M.~Dorigo, ``{Fault detection
  in autonomous robots based on fault injection and learning},''
  \emph{Autonomous Robots}, vol.~24, no.~1, pp. 49--67, 2008.

\bibitem{millard2016exogenous}
A.~G. Millard, ``{Exogenous fault detection in swarm robotic systems: an
  approach based on internal models},'' PhD Thesis, University of York, 2016.

\bibitem{Christensen2009}
A.~L. Christensen, R.~O. Grady, and M.~Dorigo, ``{From Fireflies to
  Fault-Tolerant Swarms of Robots},'' \emph{IEEE Transactions on Evolutionary
  Computation}, vol.~13, no.~4, pp. 754--766, 2009.

\bibitem{Tarapore2017}
D.~Tarapore, A.~L. Christensen, and J.~Timmis, ``{Generic , scalable and
  decentralized fault detection for robot swarms},'' \emph{PLoS ONE}, vol.~12,
  no.~8, pp. 1--29, 2017.

\bibitem{Yang2018}
G.~Z. Yang, J.~Bellingham, P.~E. Dupont, P.~Fischer, L.~Floridi, R.~Full,
  N.~Jacobstein, V.~Kumar, M.~McNutt, R.~Merrifield, B.~J. Nelson,
  B.~Scassellati, M.~Taddeo, R.~Taylor, M.~Veloso, Z.~L. Wang, and R.~Wood,
  ``{The grand challenges of Science Robotics},'' \emph{Science Robotics},
  vol.~3, no.~14, 2018.

\bibitem{chatzilygeroudis2018survey}
K.~Chatzilygeroudis and V.~Vassiliades, ``{A survey on policy search algorithms
  for learning robot controllers in a handful of trials},'' \emph{arXiv
  preprint arXiv:1807.02303v4}, pp. 1--21, 2019.

\bibitem{Bongard2006}
J.~Bongard, V.~Zykov, and H.~Lipson, ``{Resilient Machines Through Continuous
  Self-Modeling},'' \emph{Science}, vol. 314, no. November, 2006.

\bibitem{Koos2013b}
S.~Koos, A.~Cully, and J.~B. Mouret, ``{Fast damage recovery in robotics with
  the T-resilience algorithm},'' \emph{International Journal of Robotics
  Research}, vol.~32, no.~14, pp. 1700--1723, 2013.

\bibitem{Chatzilygeroudis2018a}
\BIBentryALTinterwordspacing
K.~Chatzilygeroudis, V.~Vassiliades, and J.~B. Mouret, ``{Reset-free
  Trial-and-Error Learning for Robot Damage Recovery},'' \emph{Robotics and
  Autonomous Systems}, vol. 100, pp. 236--250, 2018. [Online]. Available:
  \url{https://doi.org/10.1016/j.robot.2017.11.010}
\BIBentrySTDinterwordspacing

\bibitem{Cully2015}
A.~Cully, J.~Clune, D.~Tarapore, and J.~B. Mouret, ``{Robots that can adapt
  like animals},'' \emph{Nature}, vol. 521, no. 7553, pp. 503--507, 2015.

\bibitem{Pugh2016a}
J.~K. Pugh, L.~B. Soros, and K.~O. Stanley, ``{Quality Diversity: A New
  Frontier for Evolutionary Computation},'' \emph{Frontiers in Robotics and
  AI}, vol.~3, no. July, pp. 1--17, 2016.

\bibitem{Cully2018}
A.~Cully and Y.~Demiris, ``{Quality and Diversity Optimization: A Unifying
  Modular Framework},'' \emph{IEEE Transactions on Evolutionary Computation},
  vol.~22, no.~2, pp. 245--259, 2018.

\bibitem{Mouret2015}
J.-b. Mouret and J.~Clune, ``{Illuminating search spaces by mapping elites},''
  \emph{arXiv preprint arXiv:1504.04909v1}, pp. 1--15, 2015.

\bibitem{Pugh2015}
J.~K. Pugh, L.~B. Soros, P.~A. Szerlip, and K.~O. Stanley, ``{Confronting the
  Challenge of Quality Diversity},'' in \emph{GECCO '15: Genetic and
  Evolutionary Computation Conference}.\hskip 1em plus 0.5em minus 0.4em\relax
  New York, NY, USA: ACM, 2015.

\bibitem{Nordmoen2018}
\BIBentryALTinterwordspacing
J.~Nordmoen, K.~O. Ellefsen, and K.~Glette, ``{Combining MAP-Elites and
  Incremental Evolution to Generate Gaits for a Mammalian Quadruped Robot},''
  in \emph{21st International Conference, EvoApplications 2018}, K.~Sim and
  P.~Kaufmann, Eds.\hskip 1em plus 0.5em minus 0.4em\relax Parma, Italy:
  Springer International Publishing, 2018, pp. 159--170. [Online]. Available:
  \url{http://dx.doi.org/10.1007/978-3-319-77538-8{\_}12}
\BIBentrySTDinterwordspacing

\bibitem{Engebraten}
S.~Engebr{\aa}ten, O.~Yakimenko, J.~Moen, and K.~Glette, ``{Towards a
  Multi-Function Swarm That Adapts to User Preferences},'' in
  \emph{International Conference on Robotics and Automation (ICRA 2018)}, 2018.

\bibitem{Hart2018}
\BIBentryALTinterwordspacing
E.~Hart, A.~S. Steyven, and B.~Paechter, ``{Evolution of a functionally diverse
  swarm via a novel decentralised quality-diversity algorithm},'' in
  \emph{Proceedings of the 2018 Genetic and Evolutionary Computation Conference
  (GECCO 2018)}, Kyoto, Japan, 2018, pp. 101--108. [Online]. Available:
  \url{http://arxiv.org/abs/1804.07655{\%}0Ahttp://dx.doi.org/10.1145/3205455.3205481}
\BIBentrySTDinterwordspacing

\bibitem{Meyerson2016}
E.~Meyerson, J.~Lehman, and R.~Miikkulainen, ``{Learning Behavior
  Characterizations for Novelty Search},'' in \emph{GECCO '16: Genetic and
  Evolutionary Computation Conference}, 2016, pp. 149--156.

\bibitem{Gomez2009}
F.~J. Gomez, ``{Sustaining diversity using behavioral information distance},''
  in \emph{GECCO '09: Genetic and Evolutionary Computation Conference},
  Montr{\'{e}}al Qu{\'{e}}bec, Canada, 2009, pp. 113--120.

\bibitem{Gomes2014}
J.~Gomes, P.~Mariano, and A.~Christensen, ``{Systematic Derivation of Behaviour
  Characterisations in Evolutionary Robotics},'' in \emph{ALIFE 14: Proceedings
  of the Fourteenth International Conference on the Synthesis and Simulation of
  Living Systems}, 2014, pp. 212--219.

\bibitem{GemanS.BienenstockE.&Doursat}
S.~Geman, E.~Bienenstock, and R.~Doursat, ``{Neural networks and the bias
  variance dilemma},'' \emph{Neural Computation}, vol.~4, pp. 1--58, 1992.

\bibitem{Thompson1999}
A.~Thompson, P.~Layzell, and R.~S. Zebulum, ``{Explorations in design space:
  Unconventional electronics design through artificial evolution},'' \emph{IEEE
  Transactions on Evolutionary Computation}, vol.~3, no.~3, pp. 167--195, 1999.

\bibitem{Koos2013c}
S.~Koos, J.~B. Mouret, and S.~Doncieux, ``{The transferability approach:
  Crossing the reality gap in evolutionary robotics},'' \emph{IEEE Transactions
  on Evolutionary Computation}, vol.~17, no.~1, pp. 122--145, 2013.

\bibitem{IlgeAkkaya}
{Ilge Akkaya}, M.~Andrychowicz, M.~Chociej, M.~Litwin, B.~McGrew, A.~Petron,
  A.~Paino, M.~Plappert, G.~Powell, R.~Ribas, J.~Schneider, N.~Tezak,
  J.~Tworek, P.~Welinder, L.~Weng, Q.~Yuan, W.~Zaremba, and L.~Zhang,
  ``{Solving Rubik's Cube with a Robot Hand},'' \emph{arXiv preprint}, pp.
  1--51, 2019.

\bibitem{Pinville2011}
T.~Pinville, S.~Koos, J.~B. Mouret, and S.~Doncieux, ``{How to promote
  generalisation in evolutionary robotics: The ProGAb approach},''
  \emph{Genetic and Evolutionary Computation Conference, GECCO'11}, pp.
  259--266, 2011.

\bibitem{Schmidhuber2013}
\BIBentryALTinterwordspacing
J.~H. Schmidhuber, ``{PowerPlay: Training an Increasingly General Problem
  Solver by Continually Searching for the Simplest Still Unsolvable Problem.}''
  \emph{Frontiers in psychology}, vol.~4, no. June, p. 313, 2013. [Online].
  Available:
  \url{http://www.pubmedcentral.nih.gov/articlerender.fcgi?artid=3675324{\&}tool=pmcentrez{\&}rendertype=abstract}
\BIBentrySTDinterwordspacing

\bibitem{Srivastava2012}
R.~K. Srivastava, B.~R. Steunebrink, M.~Stollenga, and J.~Schmidhuber,
  ``{Continually adding self-invented problems to the repertoire: First
  experiments with POWERPLAY},'' \emph{2012 IEEE International Conference on
  Development and Learning and Epigenetic Robotics, ICDL 2012}, 2012.

\bibitem{Wang2019}
\BIBentryALTinterwordspacing
R.~Wang, J.~Lehman, J.~Clune, and K.~O. Stanley, ``{Paired Open-Ended
  Trailblazer (POET): Endlessly Generating Increasingly Complex and Diverse
  Learning Environments and Their Solutions},'' \emph{arXiv preprint}, pp.
  1--28, 2019. [Online]. Available: \url{http://arxiv.org/abs/1901.01753}
\BIBentrySTDinterwordspacing

\bibitem{Brant2017}
J.~C. Brant and K.~O. Stanley, ``{Minimal criterion coevolution: A new approach
  to open-ended search},'' \emph{GECCO 2017 - Proceedings of the 2017 Genetic
  and Evolutionary Computation Conference}, no. Gecco, pp. 67--74, 2017.

\bibitem{Nguyen2016}
A.~Nguyen, J.~Yosinski, and J.~Clune, ``{Innovation engines: Automated
  creativity and improved stochastic optimization via deep learning},'' in
  \emph{Proceedings of the Genetic and Evolutionary Computation Conference
  (GECCO 2015).}, 2015, pp. 545--572.

\bibitem{Huizinga2018}
\BIBentryALTinterwordspacing
J.~Huizinga and J.~Clune, ``{Evolving Multimodal Robot Behavior via Many
  Stepping Stones with the Combinatorial Multi-Objective Evolutionary
  Algorithm},'' \emph{arXiv preprint}, pp. 1--21, 2018. [Online]. Available:
  \url{http://arxiv.org/abs/1807.03392}
\BIBentrySTDinterwordspacing

\bibitem{Vassiliades2018c}
\BIBentryALTinterwordspacing
V.~Vassiliades and J.-b. Mouret, ``{Discovering the Elite Hypervolume by
  Leveraging Interspecies Correlation},'' in \emph{Proceedings of the 2018
  Genetic and Evolutionary Computation Conference (GECCO 2018)}, Kyoto, Japan,
  2018, pp. 623--630. [Online]. Available:
  \url{http://arxiv.org/abs/1804.03906{\%}0Ahttp://dx.doi.org/10.1145/3205455.3205602}
\BIBentrySTDinterwordspacing

\bibitem{TaraporeClune2016}
D.~Tarapore, J.~Clune, A.~Cully, and J.-b. Mouret, ``{How Do Different
  Encodings Influence the Performance of the MAP-Elites Algorithm ?}'' in
  \emph{Proceedings of the 2016 Genetic and Evolutionary Computation Conference
  (GECCO '16)}, Denver, CO, USA, 2016, pp. 173--180.

\bibitem{Mouret2015a}
\BIBentryALTinterwordspacing
N.~Justesen, S.~Risi, and J.-B. Mouret, ``{MAP-Elites for Noisy Domains by
  Adaptive Sampling},'' in \emph{Proceeding of the 2019 Genetic and
  Evolutionary Computation Conference (GECCO '19)}, 2019, pp. 121--122.
  [Online]. Available: \url{http://arxiv.org/abs/1504.04909}
\BIBentrySTDinterwordspacing

\bibitem{Pinciroli2012}
C.~Pinciroli, V.~Trianni, R.~O'Grady, G.~Pini, A.~Brutschy, M.~Brambilla,
  N.~Mathews, E.~Ferrante, G.~{Di Caro}, F.~Ducatelle, M.~Birattari, L.~M.
  Gambardella, and M.~Dorigo, ``{ARGoS: A modular, parallel, multi-engine
  simulator for multi-robot systems},'' \emph{Swarm Intelligence}, vol.~6,
  no.~4, pp. 271--295, 2012.

\bibitem{Riedo2013}
F.~Riedo, M.~Chevalier, S.~Magnenat, and F.~Mondada, ``{Thymio II, a robot that
  grows wiser with children},'' \emph{Proceedings of IEEE Workshop on Advanced
  Robotics and its Social Impacts, ARSO}, no. June 2016, pp. 187--193, 2013.

\bibitem{Deb2001}
\BIBentryALTinterwordspacing
K.~Deb, \emph{{Multi-Objective Optimization using Evolutionary
  Algorithms}}.\hskip 1em plus 0.5em minus 0.4em\relax Wiley, 2001. [Online].
  Available: \url{http://sutlib2.sut.ac.th/sut{\_}contents/H129518.pdf}
\BIBentrySTDinterwordspacing

\bibitem{chamanbaz2017swarm}
M.~Chamanbaz, D.~Mateo, B.~M. Zoss, G.~Toki{\'c}, E.~Wilhelm, R.~Bouffanais,
  and D.~K. Yue, ``Swarm-enabling technology for multi-robot systems,''
  \emph{Frontiers in Robotics and AI}, vol.~4, p.~12, 2017.

\bibitem{Gomes2018a}
\BIBentryALTinterwordspacing
J.~Gomes and A.~L. Christensen, ``{Task-Agnostic Evolution of Diverse
  Repertoires of Swarm Behaviours},'' in \emph{11th conference on Swarm
  Intelligence}.\hskip 1em plus 0.5em minus 0.4em\relax Springer International
  Publishing, 2018, pp. 225--238. [Online]. Available:
  \url{http://link.springer.com/10.1007/978-3-030-00533-7}
\BIBentrySTDinterwordspacing

\bibitem{Vassiliades2018d}
V.~Vassiliades, K.~Chatzilygeroudis, and J.~B. Mouret, ``{Using Centroidal
  Voronoi Tessellations to Scale Up the Multidimensional Archive of Phenotypic
  Elites Algorithm},'' \emph{IEEE Transactions on Evolutionary Computation},
  vol.~22, no.~4, pp. 623--630, 2018.

\bibitem{Cliff}
N.~Cliff, ``{Dominance statistics: Ordinal analyses to answer ordinal
  questions},'' \emph{Psychological Bulletin}, vol. 114, no.~3, pp. 494--509,
  1993.

\bibitem{Vargha2000}
A.~Vargha and H.~D. Delaney, ``{A critique and improvement of the CL common
  language effect size statistics of McGraw and Wong},'' \emph{Journal of
  Educational and Behavioral Statistics}, vol.~25, no.~2, pp. 101--132, 2000.

\bibitem{MouretJ.-B.andDoncieux2010}
S.~{Mouret, J.-B. and Doncieux}, ``{SFERESv2: Evolvin' in the Multi-Core
  World},'' in \emph{Proceedings of Congress on Evolutionary Computation
  (CEC)}, 2010, pp. 4079--4086.

\bibitem{Tarapore2019}
D.~Tarapore, J.~Timmis, and A.~L. Christensen, ``{Fault Detection in a Swarm of
  Physical Robots Based on Behavioral Outlier Detection},'' \emph{IEEE
  Transactions on Robotics}, pp. 1--7, 2019.

\bibitem{Kistemaker2011}
S.~Kistemaker and S.~Whiteson, ``{Critical factors in the performance of
  novelty search},'' in \emph{Proceedings of the Genetic and Evolutionary
  Computation Conference (GECCO 2011)}, 2011, pp. 965--972.

\bibitem{Pugh2016}
J.~K. Pugh, L.~Soros, and K.~O. Stanley, ``{Searching for Quality Diversity
  When Diversity is Unaligned with Quality},'' in \emph{PPSN 2016: Parallel
  Problem Solving from Nature – PPSN XIV}, 2016, pp. 880--889.

\bibitem{Shahriari2016}
B.~Shahriari, K.~Swersky, Z.~Wang, R.~P. Adams, and N.~{De Freitas}, ``{Taking
  the human out of the loop: A review of Bayesian optimization},''
  \emph{Proceedings of the IEEE}, vol. 104, no.~1, pp. 148--175, 2016.

\bibitem{Ligot2019}
\BIBentryALTinterwordspacing
A.~Ligot and M.~Birattari, ``{Simulation-only experiments to mimic the effects
  of the reality gap in the automatic design of robot swarms},'' \emph{Swarm
  Intelligence}, 2019. [Online]. Available:
  \url{https://doi.org/10.1007/s11721-019-00175-w}
\BIBentrySTDinterwordspacing

\bibitem{LehmanStanley2011}
J.~Lehman and K.~O. Stanley, ``{Evolving a Diversity of Creatures through
  Novelty Search and Local Competition},'' in \emph{Proceedings of the Genetic
  and Evolutionary Computation Conference (GECCO 2011)}.\hskip 1em plus 0.5em
  minus 0.4em\relax Dublin, Ireland: ACM, New York, 2011, p. 1408.

\end{thebibliography}


\begin{thebibliography}{1}
\providecommand{\url}[1]{#1}
\csname url@rmstyle\endcsname
\providecommand{\newblock}{\relax}
\providecommand{\bibinfo}[2]{#2}
\providecommand\BIBentrySTDinterwordspacing{\spaceskip=0pt\relax}
\providecommand\BIBentryALTinterwordstretchfactor{4}
\providecommand\BIBentryALTinterwordspacing{\spaceskip=\fontdimen2\font plus
\BIBentryALTinterwordstretchfactor\fontdimen3\font minus
  \fontdimen4\font\relax}
\providecommand\BIBforeignlanguage[2]{{%
\expandafter\ifx\csname l@#1\endcsname\relax
\typeout{** WARNING: IEEEtran.bst: No hyphenation pattern has been}%
\typeout{** loaded for the language `#1'. Using the pattern for}%
\typeout{** the default language instead.}%
\else
\language=\csname l@#1\endcsname
\fi
#2}}

\bibitem{Vargha2000}
A.~Vargha and H.~D. Delaney, ``{A critique and improvement of the CL common
  language effect size statistics of McGraw and Wong},'' \emph{Journal of
  Educational and Behavioral Statistics}, vol.~25, no.~2, pp. 101--132, 2000.

\end{thebibliography}
\bibliographystyle{IEEEtran}

\end{document}


\thispagestyle{empty}
\pagestyle{empty}

\maketitle


\begin{figure}[htbp!]
\centering
\includegraphics[width=1.0\textwidth]{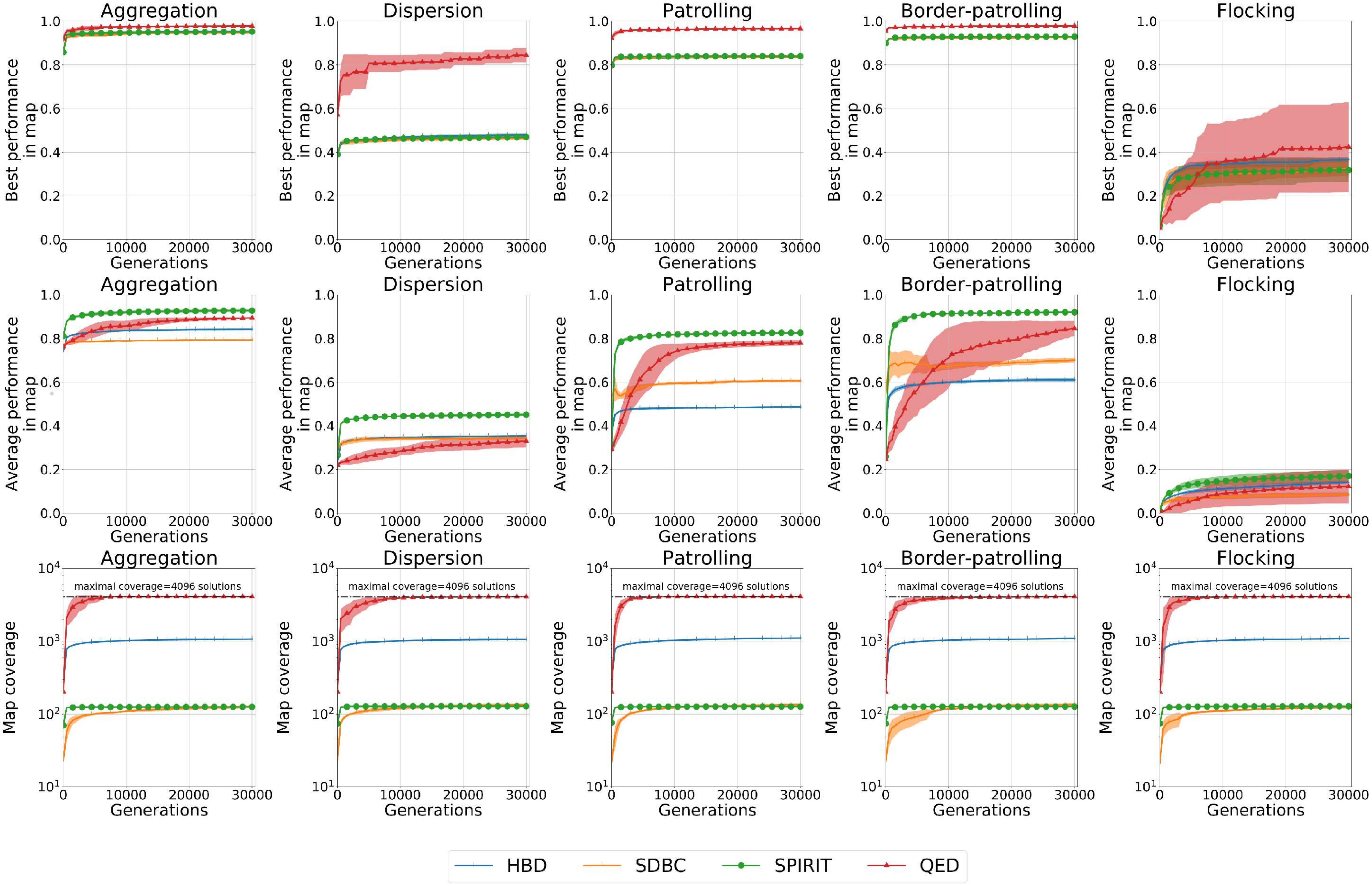} 
\caption{Evolution of best performance, average performance and map coverage (Mean $\pm$ SD) over generations, with solid line indicating the Mean across replicates while the shaded area represents the Mean $\pm$ SD. For each algorithm, data are based on 5 independent replicates per task. The performance of a solution is the average fitness across 50 independent trials. The best and average performance of a map are defined as the maximum and mean performance across its solutions. The coverage of a map is the proportion of cells it has filled with non-empty solutions. } \label{fig: evolution}
\end{figure}

\begin{table*}[htb!]
\centering
\caption{Summary statistics of the best performance, average performance, and the coverage (Mean $\pm$ SD), all evaluated at the end of the evolutionary replicates, with Mean and SD being taken across 5 different replicates. Performance of a solution is defined as the average fitness across 10 independent trials, and is expressed as a proportion of its empirical maximum across all replicates and all algorithms for a given task. The best and average performance of a map are defined as the maximum and mean performance across its solutions. The coverage of a map is the proportion of cells it has filled with non-empty solutions.} \label{tab: evolutionstats}
\resizebox{1.0\textwidth}{!}{
\begin{tabular}{l | p{1.53cm} p{1.53cm} p{1.53cm} p{1.53cm} p{1.53cm} p{1.53cm} p{1.53cm}  p{1.53cm} p{1.53cm} p{1.53cm}  p{1.53cm} p{1.53cm} p{1.53cm} p{1.53cm}}
& \multicolumn{12}{l}{\textbf{Condition}}\\ \hline
\textbf{Swarm task}    & \multicolumn{3}{c}{HBD}     & \multicolumn{3}{c}{SDBC}     & \multicolumn{3}{c}{SPIRIT}     & \multicolumn{3}{c}{QED}  \\ \hline
 & Best \newline performance  & Average \newline performance  & Coverage  & Best \newline performance  & Average \newline performance  & Coverage  & Best \newline performance  & Average \newline performance  & Coverage  & Best \newline performance  & Average \newline performance  & Coverage \\ 
Aggregation & $0.99 \pm 0.00$  & $0.87 \pm 0.01$  & $0.26 \pm 0.00$  & $0.99 \pm 0.01$  & $0.81 \pm 0.00$  & $0.03 \pm 0.00$  & $0.99 \pm 0.01$  & $0.96 \pm 0.01$  & $0.03 \pm 0.00$  & $0.98 \pm 0.00$  & $0.91 \pm 0.01$  & $1.00 \pm 0.00$ \\ 
Dispersion & $0.98 \pm 0.01$  & $0.72 \pm 0.00$  & $0.26 \pm 0.01$  & $0.95 \pm 0.01$  & $0.69 \pm 0.01$  & $0.03 \pm 0.00$  & $0.96 \pm 0.02$  & $0.92 \pm 0.01$  & $0.03 \pm 0.00$  & $0.84 \pm 0.02$  & $0.60 \pm 0.04$  & $1.00 \pm 0.00$ \\ 
Patrolling & $0.99 \pm 0.00$  & $0.58 \pm 0.00$  & $0.27 \pm 0.00$  & $0.99 \pm 0.00$  & $0.71 \pm 0.01$  & $0.03 \pm 0.00$  & $1.00 \pm 0.00$  & $0.98 \pm 0.00$  & $0.03 \pm 0.00$  & $0.98 \pm 0.01$  & $0.92 \pm 0.03$  & $1.00 \pm 0.00$ \\ 
Border-patrolling & $1.00 \pm 0.00$  & $0.65 \pm 0.01$  & $0.27 \pm 0.00$  & $0.99 \pm 0.00$  & $0.74 \pm 0.01$  & $0.03 \pm 0.00$  & $1.00 \pm 0.00$  & $0.99 \pm 0.00$  & $0.03 \pm 0.00$  & $0.98 \pm 0.00$  & $0.90 \pm 0.05$  & $1.00 \pm 0.00$ \\ 
Flocking & $0.99 \pm 0.01$  & $0.36 \pm 0.01$  & $0.27 \pm 0.01$  & $0.85 \pm 0.11$  & $0.21 \pm 0.02$  & $0.03 \pm 0.00$  & $0.84 \pm 0.14$  & $0.44 \pm 0.07$  & $0.03 \pm 0.00$  & $0.60 \pm 0.22$  & $0.24 \pm 0.15$  & $1.00 \pm 0.00$ \\ 

\end{tabular}
}
\end{table*}

\begin{table*}[htb!]
\centering
\caption{Summary statistics of fault recovery scores (Median $\pm$ IQR) aggregated across all faults and all tasks. Performance is obtained by averaging the fitness of a solution across 10 independent trials, and is expressed as a proportion of its empirical maximum across all replicates and all algorithms for a given task.  Impact of the fault is the drop in performance when the best controller for the normal operating environment is evaluated in the faulty environments. Recovered performance is the performance of the recovery solution in the faulty environments. Resilience is the proportional drop in performance after recovery. The data are based on 5 tasks, 5 unique maps from the different evolutionary replicates, and 50 randomly sampled faults to each archive, yielding 1250 data points per task per algorithm.} \label{tab: significance}
\begin{tabular}{l | p{2cm} p{2cm} p{2cm} } 
 & Impact of \newline fault   &  Recovered \newline performance  & Resilience \\ \hline
HBD & $-0.25 \pm 0.33$  & $0.87 \pm 0.11$  & $-0.12 \pm 0.11$ \\ 
SDBC & $-0.24 \pm 0.26$  & $0.87 \pm 0.10$  & $-0.11 \pm 0.08$ \\ 
SPIRIT & $-0.25 \pm 0.29$  & $0.86 \pm 0.13$  & $-0.12 \pm 0.13$ \\ 
QED & $-0.23 \pm 0.26$  & $0.88 \pm 0.16$  & $-0.08 \pm 0.07$ \\ 
\end{tabular}
\end{table*}

\begin{table*}[htb!]
\centering
\caption{Significance data comparing QED's resilience and recovered performance scores (Median $\pm$ IQR) to the other quality-diversity algorithms for each task. Performance is obtained by averaging the fitness of a solution across 10 independent trials, and is expressed as a proportion of its empirical maximum across all replicates and all algorithms for a given task.  Resilience is the proportional drop in performance after recovery and recovered performance is the performance of the recovery solution. The data are based on 5 unique maps from the different evolutionary replicates and 50 randomly sampled faults to each archive, yielding 250 data points per task per algorithm. Significance is tested using the Wilcoxon rank-sum test and the effect size metric is Cliff's $\delta$. Superscripts $^{*}$ and $^{**}$ indicate significance with significance criterion $\alpha=0.05/m$ and $\alpha=0.001/m$, respectively, where $m=15$ is the number of comparisons used as Bonferroni correction. Bold font indicates large effect size, defined as $|\delta| \geq .43$ \cite{Vargha2000}, and the sign indicates the direction of the effect (positive if QED outperforms the other algorithm; negative otherwise).} \label{tab: significance}

\subfloat[Resilience]{ \label{tab: resilience-sign}
\resizebox{1.0\textwidth}{!}{
\begin{tabular}{l | l | lll | lll | lll } 
& \textbf{QED}   &\multicolumn{9}{l}{\textbf{Comparison}}\\ \hline
\textbf{Swarm task}&   & \multicolumn{3}{c|}{HBD}& \multicolumn{3}{c|}{SDBC}& \multicolumn{3}{c|}{SPIRIT}\\ \hline
& Resilience & Resilience & Significance & Effect& Resilience & Significance & Effect& Resilience & Significance & Effect\\ \hline
Aggregation& $-0.05 \pm 0.0$& $-0.06 \pm 0.0$ & $p=0.396$ & $0.04$& $-0.08 \pm 0.0$ & $p<0.001^{**}$ & $0.43$& $-0.07 \pm 0.0$ & $p<0.001^{**}$ & $0.28$\\ 
Dispersion& $-0.08 \pm 0.0$& $-0.11 \pm 0.0$ & $p<0.001^{**}$ & $\mathbf{0.55}$& $-0.11 \pm 0.0$ & $p<0.001^{**}$ & $\mathbf{0.56}$& $-0.12 \pm 0.1$ & $p<0.001^{**}$ & $\mathbf{0.68}$\\ 
Patrolling& $-0.10 \pm 0.1$& $-0.14 \pm 0.1$ & $p<0.001^{**}$ & $\mathbf{0.51}$& $-0.11 \pm 0.1$ & $p<0.001^{*}$ & $0.17$& $-0.15 \pm 0.1$ & $p<0.001^{**}$ & $\mathbf{0.60}$\\ 
Border-patrolling & $-0.06 \pm 0.0$& $-0.11 \pm 0.0$ & $p<0.001^{**}$ & $\mathbf{0.72}$& $-0.08 \pm 0.0$ & $p<0.001^{**}$ & $\mathbf{0.49}$& $-0.10 \pm 0.1$ & $p<0.001^{**}$ & $\mathbf{0.63}$\\ 
Flocking& $-0.64 \pm 0.3$& $-0.76 \pm 0.1$ & $p<0.001^{**}$ & $\mathbf{0.44}$& $-0.76 \pm 0.2$ & $p<0.001^{**}$ & $0.40$& $-0.79 \pm 0.2$ & $p<0.001^{**}$ & $\mathbf{0.55}$\\ 
\end{tabular}
}
}

\subfloat[Recovered performance]{ \label{tab: performance-sign}
\resizebox{0.99\textwidth}{!}{
\begin{tabular}{l | p{1.8cm} | p{1.8cm}ll | p{1.8cm}ll | p{1.8cm}ll } 
& \textbf{QED}   &\multicolumn{9}{l}{\textbf{Comparison}}\\ \hline
\textbf{Swarm task}&   & \multicolumn{3}{c|}{HBD}& \multicolumn{3}{c|}{SDBC}& \multicolumn{3}{c|}{SPIRIT}\\ \hline
& Recovered \newline performance & Recovered \newline performance & Significance & Effect& Recovered  \newline performance & Significance & Effect& Recovered \newline performance & Significance & Effect\\ \hline
Aggregation& $0.93 \pm 0.0$& $0.94 \pm 0.0$ & $p<0.001^{*}$ & $-0.18$& $0.91 \pm 0.0$ & $p<0.001^{**}$ & $0.35$& $0.93 \pm 0.0$ & $p=0.135$ & $0.08$\\ 
Dispersion& $0.77 \pm 0.0$& $0.87 \pm 0.0$ & $p<0.001^{**}$ & $\mathbf{-0.98}$& $0.85 \pm 0.0$ & $p<0.001^{**}$ & $\mathbf{-0.88}$& $0.84 \pm 0.1$ & $p<0.001^{**}$ & $\mathbf{-0.81}$\\ 
Patrolling& $0.89 \pm 0.1$& $0.85 \pm 0.1$ & $p<0.001^{**}$ & $0.42$& $0.88 \pm 0.1$ & $p=0.895$ & $0.01$& $0.85 \pm 0.1$ & $p<0.001^{**}$ & $\mathbf{0.44}$\\ 
Border-patrolling& $0.92 \pm 0.0$& $0.89 \pm 0.0$ & $p<0.001^{**}$ & $\mathbf{0.50}$& $0.91 \pm 0.0$ & $p<0.001^{**}$ & $0.26$& $0.90 \pm 0.0$ & $p<0.001^{**}$ & $0.38$\\ 
Flocking& $0.21 \pm 0.1$& $0.24 \pm 0.1$ & $p=0.003^{*}$ & $-0.15$& $0.20 \pm 0.1$ & $p=0.823$ & $-0.01$& $0.17 \pm 0.1$ & $p<0.001^{**}$ & $0.22$\\ 
\end{tabular}
}
}
\end{table*}
\clearpage


\bibliography{library} 
\bibliographystyle{IEEEtran}